\documentclass[11pt]{article}
\usepackage[preprint]{acl}
\usepackage{times}
\usepackage{latexsym}
\usepackage[T1]{fontenc}
\usepackage[utf8]{inputenc}
\usepackage{microtype}
\usepackage{inconsolata}
\usepackage{graphicx}
\usepackage{subcaption}
\usepackage{listings}
\usepackage[most]{tcolorbox}
\usepackage{array}
\usepackage{booktabs,multirow}
\usepackage{siunitx}
\usepackage{tikz}
\usepackage{pgfplots}
\pgfplotsset{compat=1.18}
\usepgfplotslibrary{groupplots}
\usepackage{enumitem}
\usepackage{amssymb}
\usetikzlibrary{decorations.pathreplacing}
\usepackage{pifont}
\usepackage{xcolor}
\usepackage{hyperref}

\lstdefinestyle{spicecompact}{
  basicstyle=\ttfamily\tiny,
  frame=single,
  breaklines=true,
  columns=fullflexible,
  aboveskip=0pt,
  belowskip=0pt
}

\newcommand{\update}[1]{\textcolor{black}{#1}}

\newcommand{\systemname}{\textsc{Sparc}} 
\newcommand{\D}{\textsf{D}}
\newcommand{\N}{\textsf{N}}
\newcommand{\Q}{\textsf{Q}}
\newcommand{\Po}{\textsf{P}}

\newcommand{\amsnet}{NetQ}
\newcommand{\eeebench}{CktBench}

\newenvironment{packeditemize}{
\begin{itemize}[leftmargin=*]
  \setlength{\itemsep}{0pt}
  \setlength{\parskip}{0pt}
  \setlength{\parsep}{0pt}
}{\end{itemize}}

\title{\systemname{}: A Multi-Agent System for Electrical Circuit Question Answering}
\author{
  \textbf{Mushtari Sadia$^+$} \quad
  \textbf{Zhenning Yang$^+$} \quad
  \textbf{Umme Habiba Lamia$^*$} \quad
  \textbf{Nishat Shawrin$^*$} \\
  \textbf{Ang Chen$^+$} \quad
  \textbf{Amrita Roy Chowdhury$^+$} \\
  $^+$University of Michigan, Ann Arbor, $^*$Bangladesh University of Engineering and Technology
}

\begin{document}
\maketitle
\begin{abstract}
Electrical circuit diagram QA tasks require complex mathematical reasoning, which remains challenging for multimodal LLMs. We present \systemname{}, a multi-agent system that answers questions over circuit diagrams by grounding reasoning in executable physics-based simulations. \systemname{} uses LLM agents to synthesize, execute, and analyze simulation programs, improving accuracy and reliability by design. It achieves 83\% accuracy, with up to a 58\% absolute improvement over baselines, while enabling systematic error diagnosis. \footnote{Both our datasets and code are publicly available at: \href{https://github.com/Mushtari-Sadia/sparc.git}{https://github.com/Mushtari-Sadia/sparc.git}}
\end{abstract}
\section{Introduction}
\label{sec:intro}

Electrical circuit diagrams are ubiquitous in engineering, serving as the primary medium through which engineers communicate and reason about electrical systems. Answering questions about these diagrams (e.g., predicting behavior, checking correctness, or resolving design queries) typically requires domain experts with deep knowledge of electrical engineering and physics. This reliance on expert reasoning creates a major practical bottleneck. Automating natural language question answering over circuit diagrams would therefore enable faster design cycles, scalable verification, and broader accessibility to engineering tools.

While recent work~\cite{li2025eee} has taken initial steps by introducing benchmark datasets, progress remains limited. Prior work, and our experiments (Sec.~\ref{sec:eval}), show that state-of-the-art models achieve only \textasciitilde{}51\% accuracy on this task. This sharply contrasts with the strong performance ($>$80\%) of multimodal LLMs on diagram-based question answering more broadly \cite{goyal2017makingvvqamatter, marino2019ok, yue2025mmmu}, highlighting that electrical circuit diagrams remain a uniquely difficult and largely unsolved challenge.

\begin{figure*}
    \centering
    \includegraphics[width=1\linewidth]{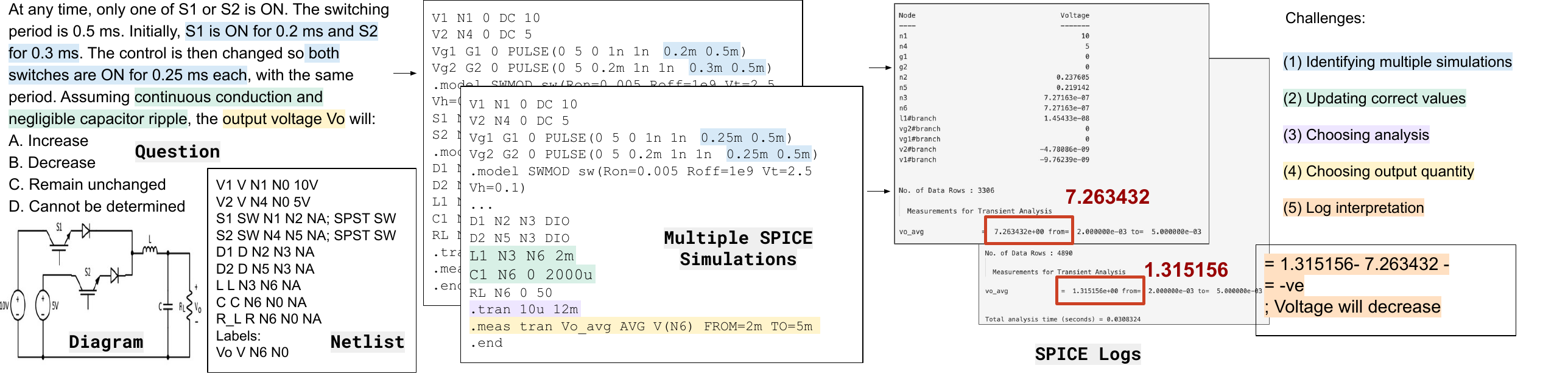}
    \caption{Challenges in circuit QA with SPICE: Given a diagram, a netlist and a question, the system must (1) determine number of simulations required; (2) construct or update a SPICE program with correct component values, possibly inferred from text; (3) select analyses (4) select measurements; and (5) interpret outputs to produce answer.}
    \label{fig:example} 
    % \vspace{-0.2cm}
\end{figure*}

At its core, circuit question answering involves mathematical reasoning, but with a key difference from traditional math problems. For example, consider the question in Fig.~\ref{fig:example}: \textit{How does the output voltage change when the switching duty cycle of a power converter is modified?}. Unlike standard math problems, the required equations are \textit{not} given and must instead be \textit{derived} from the circuit diagram. This requires identifying the relevant subcircuit, reasoning about component connectivity, selecting the appropriate physical laws (e.g., Ohm’s law), and then formulating and solving the corresponding equations. Achieving high accuracy therefore requires a multi-step process integrating perception (recognizing symbols, components), structural reasoning (topology), and domain knowledge (applying electrical laws).

Equally important, real-world engineering practice demands not only accuracy but also \textit{reliability}. In safety-critical workflows, engineers must understand why an answer is correct to validate designs, debug failures, and ensure safety. However, the probabilistic nature of LLMs means even correct answers provide no guarantee of correct reasoning. Prompting LLMs to generate explanations does not solve this problem, since verifying their mathematical validity still requires domain expertise, undermining the benefits of automation.

Meeting these \textit{dual} requirements of high accuracy and reliability in circuit QA is precisely the challenge we address. Rather than using LLMs to solve problems end-to-end, we offload mathematical reasoning to mature, domain-specific tools. Concretely, we use SPICE~\cite{nagel1975spice2}, the de facto framework for circuit simulation. A SPICE program specifies circuit components, connectivity, and analysis types, after which a simulator (e.g., \texttt{ngspice}) automatically formulates and solves the governing equations. Our key idea is to use LLMs as an \textit{interface} to SPICE: the LLM translates questions and circuit diagrams into SPICE specifications, while SPICE performs the underlying computation. As a symbolic tool purpose-built for circuit analysis, SPICE improves accuracy and provides an auditable execution trace for debugging and verification. Together, these choices enable high accuracy and reliability \textit{by design}.
%\arc{Need to say that this work is orthogonal nevertheless we include baselines}

However, operationalizing this idea is far from trivial. SPICE accepts only rigid input: (1) the circuit diagram must be translated into a netlist, a textual representation of circuit topology~\cite{nagel1975spice2}, and (2) the analysis must be expressed as an executable program. \update{The former step depends heavily on the visual perception capability of vision language models (VLMs). Recent work has already advanced automated netlist extraction, with modern VLMs achieving strong performance on benchmark datasets~\cite{shi2025amsnet, uzair2023automated}. As a fundamentally visual perception problem, netlist extraction is orthogonal to the core challenges addressed in this work. Nevertheless, to evaluate the fully end to end setting, we additionally include baselines using automatically extracted netlists (Sec.~\ref{sec:eval:ablations}, Table~\ref{tab:ablation1}). However, the latter step, synthesizing executable SPICE programs that correctly operationalize question semantics, has \textit{not} been explored in prior research and introduces substantial technical challenges.} First, the mapping from a question to a SPICE simulation is rarely one-to-one, as a single query often requires multiple simulations. Second, SPICE programs contain many numerical values, device parameters, and analysis constraints that must be populated accurately from both the diagram and the question. Third, simulation outputs must be carefully post-processed to derive the final answer, including aggregating results across multiple runs and computing quantities. Crucially, since circuit QA demands strict \textit{end-to-end} correctness, as even a minor mistake at any intermediate step can invalidate the entire reasoning chain. These challenges make single-shot program generation unreliable.

% The central novelty of \systemname{} lies in framing SPICE program generation as a \textit{structured, execution-guided process} rather than single-shot generation. Specifically,

To address these challenges, we present \systemname\footnote{SPICE Program Analysis and Reasoning for Circuits}, a multi-agent system for answering natural language questions over circuit diagrams using SPICE simulations.  \systemname{} decomposes this task into three stages: simulation \textbf{setup}, \textbf{execution}, and \textbf{analysis}. In the \textbf{simulation setup} stage,  the system interprets the natural-language question and decomposes it into independent simulation tasks. In the \textbf{simulation execution} stage, specialized agents translate each task into a SPICE program, and execution failures trigger targeted repair. Finally, in the \textbf{simulation analysis} stage, \systemname{} derives the final answer from simulator outputs via tool-based calculations. In summary, our contributions are:
\vspace{-4pt}
\begin{packeditemize}
\item We introduce \systemname{}, \textbf{a multi-agent system} for electrical circuit QA that synthesizes, executes, and analyzes SPICE programs, grounding LLM reasoning in physics-based simulation and enabling reliability through verifiable computation.
\item We \textbf{introduce two datasets} pairing circuit diagrams with complex QA pairs requiring mathematical analysis, including netlist annotation and an \emph{automated} QA generation pipeline.% \update{} Bseides this Orthogonal to the core goal of our paper, these datasets also supporting future fine tuning for netlist extraction tasks. to firther imrphve the gener 
\item We show through extensive experiments that \systemname{} achieves 83\% accuracy and up to 58\% improvement over baselines. It delivers the \textbf{largest gains for weaker models}, demonstrating its design narrows the gap to stronger models.
\item We further improve reliability by introducing a \textbf{structured error taxonomy} that localizes failures to specific pipeline stages, enabling systematic diagnosis and interpretability.
\end{packeditemize}

\section{Problem Overview}
\subsection{Task} 
\label{sec:task}
%\arc{Mention the task - say why we don't focus on netlist extraction, our dataset looks like diagram, netlist, question}

Given an electrical circuit diagram and a natural language question, our goal is to produce a \textit{numerical} answer (or one derivable from numerical values). We focus on challenging questions that require solving equations derived from laws of physics (e.g., Fig.~\ref{fig:example}), rather than vision queries (e.g., “How many resistors are in the diagram?”).

To answer such questions, we use SPICE, a physics-based circuit simulation framework. As described in Sec.~\ref{sec:intro}, this entails two key sub tasks. First, the circuit diagram must be translated into a netlist (a description of circuit topology). Second, the question must be mapped to one or more SPICE programs, followed by analysis of the simulation outputs. \update{The first subtask is largely a visual extraction problem and is therefore orthogonal to the core challenges addressed in this work. Recent methods achieve up to 90.19 F1 in recovering accurate netlists from circuit images on the AMSNet2.0 dataset~\cite{shi2025amsnet}. As modern vision models continue to improve, we expect this capability to strengthen further, a trend also reflected in our experimental results (Appendix~\ref{app:model_progression}). Additionally, the datasets we introduce provide high-quality circuit diagram--netlist pairs that could support future fine-tuning efforts for even more accurate netlist extraction.} In contrast, the second sub task is the core challenge. Synthesizing a SPICE program is \textit{not} merely a matter of syntactic correctness; it must faithfully encode the semantic intent of the question to invoke the correct simulation. We elaborate on the technical challenges in Sec.~\ref{sec:challenges}. Moreover, as our experiments show (Sec.~\ref{sec:eval}), performance on this step largely determines overall task accuracy, making it the primary bottleneck. We therefore focus on this underexplored challenge and propose \systemname{}, a multi-agent system for synthesizing SPICE programs from natural language specifications and analyzing their outputs. To the best of our knowledge, this is the first work to address this problem. Each data point in this task consists of a circuit diagram $\D$, netlist $\N$, and natural language question $\Q$.\\
\noindent \textbf{Note.} \update{We exclude questions that cannot be answered through circuit simulation, including: (1)~purely conceptual questions requiring no simulation, (2)~questions requiring real-world device parameters absent from the netlist (e.g., datasheet values), and (3)~open-ended design or optimization queries beyond SPICE. 
\\Also, as explained above, netlist extraction is a purely visual task and is orthogonal to our core contribution. Nevertheless, we include baselines evaluating the fully end to end setting using automatically extracted netlists (Sec.~\ref{sec:eval:ablations}, Table~\ref{tab:ablation1}).}

\subsection{SPICE Background}
% \arc{we should add back some of the details here because readers don;t have background in SPICE and some more explanation is necessary}
\label{sec:spice_background}
A SPICE program has the following three sections. \\
\noindent\textbf{(1) Circuit Specification.} The circuit specification defines the circuit topology and its components. \textit{Each} electrical component (e.g., resistors, batteries) must be instantiated with \textit{valid numerical} parameters, such as a battery’s voltage. 

\noindent\textbf{(2) Analysis Specification.} The next step is to select the required mathematical analyses which determines three aspects:  (1) current type--DC or AC (direct or alternating current), (2) whether transient analysis is needed to measure time-varying quantities, and (3) whether parameter sweeps are required to evaluate multiple operating conditions. 

\noindent\textbf{(3) Output Specification.} Executing the analysis produces detailed simulation traces with measurements, such as node voltages and currents. The output specification determines which values are extracted from these traces to compute the final answer, which  may require selecting values at specific time points, aggregating over time, or combining results from multiple  runs. (Details in App.~\ref{app:SPICE}).

% --- Full-width top figure ---
\begin{figure*}[t]
    \centering
    \includegraphics[width=\linewidth]{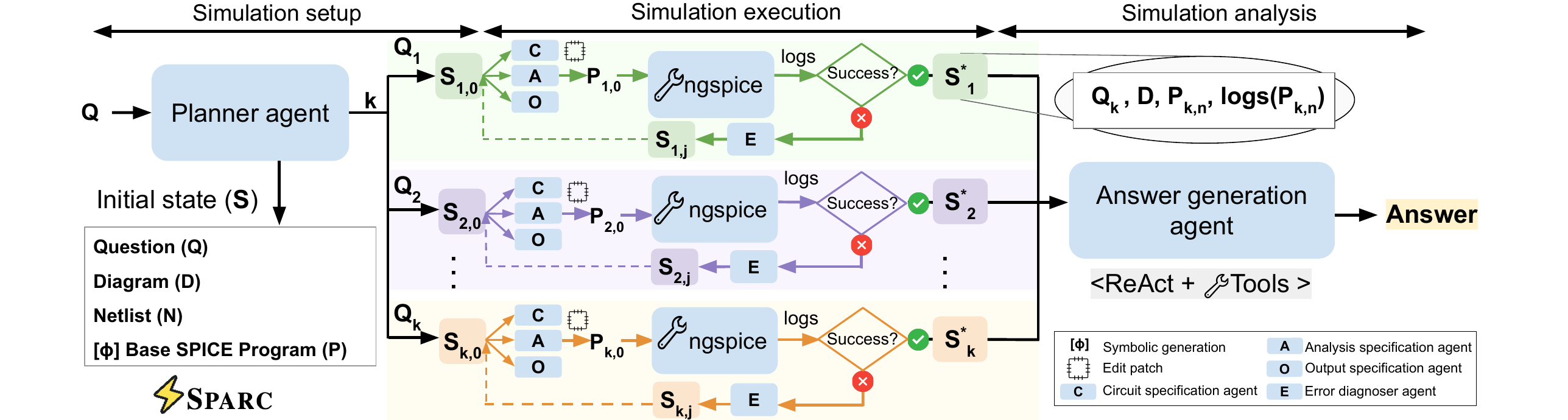}
    % \vspace{-0.8cm}
    \caption{\textbf{\systemname{} Overview.} In the \textbf{simulation setup} stage, the planner agent constructs an initial state \( \textsf{S} \) containing the circuit diagram \( \D \), a netlist \( \N \), a natural-language question \( \Q \), and a base SPICE program \( \Po \). It then analyzes \( \Q \) to determine the number of simulations \( k \), producing simulation-specific states \( \{\textsf{S}_{i,0}\} \). In the \textbf{simulation execution} stage, each simulation proceeds independently by iteratively constructing an executable SPICE program \( \Po_i \) through patch generation and repair using three agents, followed by execution with \texttt{ngspice} (an open-source SPICE circuit simulator), and updating the state with output logs from \( \Po_i \). Upon successful execution of all simulations (denoted by $\textsf{S}^*_i$), the \textbf{simulation analysis} stage post-processes the $k$ simulation outputs to derive the final answer. \update{\textbf{Fig.~\ref{fig:end-to-end-example} in the appendix presents a complete end-to-end execution example of \systemname{} across all stages.}}}
    \label{fig:system-diagram}
\end{figure*}

\begin{figure}[t]
    \centering
    \begin{subfigure}[t]{0.49\linewidth}
        \centering
        \includegraphics[width=\linewidth]{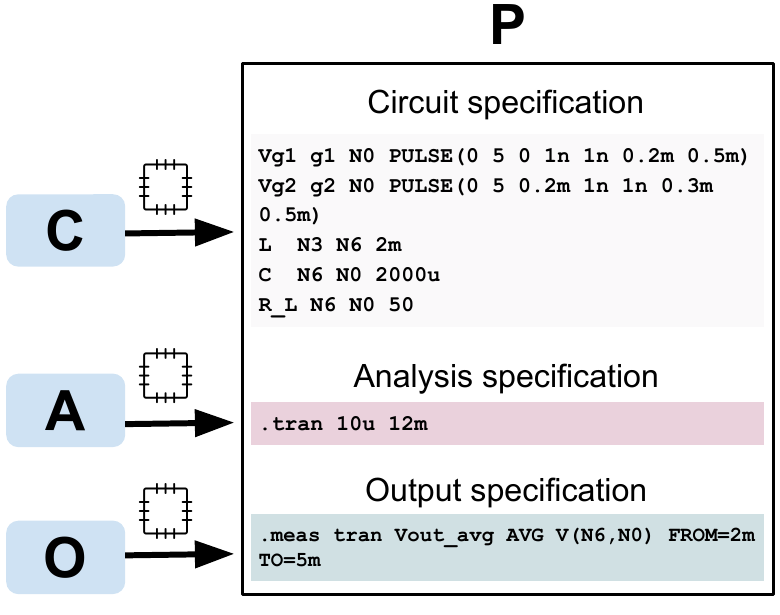}
        \caption{Section-wise patching of \Po}
        \label{fig:patch}
    \end{subfigure}
    \hfill
    \begin{subfigure}[t]{0.49\linewidth}
        \centering
        \includegraphics[width=\linewidth]{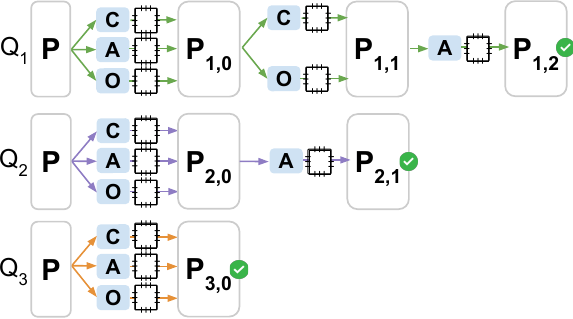}
        \caption{Repair iteration patterns}
        \label{fig:error-workflow}
    \end{subfigure}
    % \vspace{-0.3cm}
    \caption{\textbf{Execution and repair.} \textbf{(a)} Specialized agents apply patches to disjoint sections of the base program \( \Po \). \textbf{(b)} All agents are invoked in the first iteration, producing initial programs \( \Po_{i,0} \). If errors persist, only agents responsible for faulty patches are re-invoked until successful execution (e.g., only the circuit and output specification agents are invoked for \( \Po_{1,1} \)). The first simulation requires three repairs; the third requires none.}

    \label{fig:error-workflow-and-patch}
    % \vspace{-2mm}
\end{figure}

% \arc{flip the two figure order - in b) mention circuit speciations Analysis speciifican and output, somehow we need it to make it more prominent that the updates are disjoint. put more explaination for say that in the three runs one is succesfuk in firs try while the others tale more runs and not all are called sekectively called, patcvhes are applied to programs not states so use P and also use the patch symbol}
% \arc{typos: the third label should be $ P_{3,0}$ and the last of the first one should be $P_{1,2}$, Add $Q_1, Q_2, Q_3$ before the P boxes}

\vspace{-3mm}
\subsection{Technical Challenges}\label{sec:challenges}

We illustrate the technical challenges of our task using the example in Fig.~\ref{fig:example}.

\noindent\textbf{(1) Mapping Questions to SPICE Programs.}
Consider the question in Fig.~\ref{fig:example}, which asks \textit{whether the output voltage changes when the switching duty cycle is modified}. Although it appears to seek a single outcome, answering it requires multiple simulations under different switching schedules whose results must be compared. This illustrates a key challenge: many circuit questions \textit{cannot} be expressed as a single SPICE run.

Even constructing each simulation is non trivial. Circuit specification requires instantiating every component with valid parameters, which is challenging for three reasons. First, some values are missing from the diagram or netlist and must be inferred from the question (e.g., “0.2 ms” and “0.3 ms” in Fig.~\ref{fig:example}). Second, domain-specific terms must be translated into numerical constraints (e.g., “negligible ripple” implies a minimum capacitor value in Fig.~\ref{fig:example}). Third, some parameters may be missing and must be filled with reasonable defaults, such as an unspecified resistance value in Fig.~\ref{fig:example}. Beyond circuit specification, the analysis (e.g., selecting transient analysis) and output specification (e.g., averaging over a time interval) also require nontrivial semantic understanding of the question.

\noindent\textbf{(2) Execution Brittleness.} Because SPICE solves equations constrained by physics, even syntactically valid programs can fail when specifications are inconsistent. For example, in Fig.~\ref{fig:example}, incorrect switch timing can short the voltage sources by turning both switches ON simultaneously. Although the resulting program is syntactically valid, the simulator reports low level errors that obscure the root cause, requiring semantic reasoning to diagnose. Since such failures are often localized, regenerating the entire program frequently introduces errors into previously correct components.

\noindent\textbf{(3) Analyzing Simulation Outputs.}
Even when simulations execute successfully, answering the question requires analysis (e.g., Fig.~\ref{fig:example} processes two logs to determine changes in voltage).

\vspace{-1mm}

%For example, a resistor of \SI{5}{\ohm} connected between two nodes may be specified as \texttt{R1 a b 5}.
%Additionally, A SPICE program must specify how the circuit should be analyzed (e.g., simulating time varying behavior). For example, \texttt{.op} requests a DC operating point analysis, while \texttt{.tran 1u 10m} specifies a time-varying simulation. 
% For example, \texttt{.print tran V(out)} reports the output voltage during a transient analysis, while \texttt{.meas tran Vavg AVG V(out)} computes an aggregate measurement.
%The technical challenge is understanding which analysis to invoke based on the semantics of the natural-language question. 

% \vspace{-1mm}
\section{\systemname}
\vspace{-2mm}
\systemname{} is a multi-agent system for circuit question answering that operates by synthesizing, executing, and analyzing SPICE simulations. The task comprises several qualitatively different subtasks where even a small error at any stage can render the final answer invalid. \systemname{} therefore adopts a multi-agent design with specialized agents for each subtask, decomposing the process into three stages:
 \vspace{-3mm}
\begin{packeditemize}
\item \textbf{Simulation Setup}: This stage (1) initializes execution context (state), and (2) determines the number of required simulations by decomposing $\Q$ into appropriately scoped sub-questions. 
\item \textbf{Simulation Execution}: This stage (1) generates simulation-specific SPICE programs, (2) executes the generated SPICE programs, (3) diagnoses execution failures, and performs repairs.
\item \textbf{Simulation Analysis}: This stage (1) post-processes simulator outputs to extract the final answer and, when necessary, (2) performs additional mathematical reasoning.
\end{packeditemize}
% \zy{\noindent \textbf{Workflow.}
% Figure~\ref{fig:system-diagram} illustrates the \systemname{} workflow; the caption provides a step-by-step description.
% Given $(\D,\N,\Q)$, \systemname{} maintains a shared \textit{state} that stores information updated across agents and stages.
% The planner constructs an initial state $\textsf{S}$ containing $\D$, $\N$, $\Q$, and a base SPICE program $\Po$ deterministically generated from $\N$.
% It then decomposes $\Q$ into $k$ scoped sub-questions $\{\Q_i\}_{i=1}^k$ and forks $\textsf{S}$ into simulation-specific states $\{\textsf{S}_{i,0}\}_{i=1}^k$.
% Each state is processed independently by agents that patch $\Po$, execute the resulting program with ngspice, and selectively repair failures using simulator feedback.
% Once all simulations reach successful states $\{\textsf{S}_i^\ast\}_{i=1}^k$, an answer generation agent aggregates their logs to produce the final answer; otherwise, \systemname{} reports failure after a bounded number of retries.
% }
Fig.~\ref{fig:system-diagram} illustrates the \systemname{} workflow, and \update{Fig.~\ref{fig:end-to-end-example} provides a full end-to-end execution example}. Recall that each data point consists of: a circuit diagram ($\D$), its corresponding netlist ($\N$), and a natural-language question ($\Q$). \systemname{} maintains a shared memory structure called \textit{state} that stores information updated across agents and stages.

\systemname~starts with the simulation setup stage, where a \emph{planner} agent generates an initial state \textsf{S}. This state contains the data point and a base SPICE program $\Po$ which is deterministically constructed from the netlist $\N$. The planner determines how many simulations are required, and forks the initial state into simulation-specific states ${\textsf{S}_{1,0}, \ldots, \textsf{S}_{k,0}}$. Each state $\textsf{S}_{i,0}$ corresponds to a distinct simulation for a \textit{sub-question} $\Q_i$.  Next, \systemname~enters the simulation execution stage, where each of the $k$ simulations proceed \textit{independently}. For each simulation i, the goal is to construct a SPICE program $\Po_i$ by generating patches to the base program $\Po$. Mirroring the structure of a SPICE program (Sec.~\ref{sec:spice_background}), \systemname~employs three agents: a \textit{circuit specification} agent, an \textit{analysis specification} agent, and an \textit{output specification} agent, each responsible for a disjoint section of the program. These agents propose edit patches that modify only their assigned sections, yielding program $\Po_{i,0}$, which is then executed. If fails, the simulation enters an iterative repair loop. An \textit{error diagnoser} agent analyzes output logs to identify faulty sections, then selectively re-invokes only the corresponding agents to repair them. This produces refined states $\textsf{S}_{i,j}$ and programs $\Po_{i,j}$ (indexed by repair iteration $j \in \{1, \cdots, T\}$). The repair-execute cycle repeats until successful execution (or until the retry bound $T$ is reached, in which case \systemname{} terminates and reports failure). Once all simulations execute successfully, \systemname{} moves to the simulation analysis stage, where an \emph{answer generation} agent aggregates the $k$ simulation logs to produce the final answer.
\subsection{Simulation Setup}
\label{sec:setup}

% \zy{
% \noindent \textbf{Planner Agent.}
% The planner initializes the state by constructing contextual metadata for $\Q$ and a base SPICE program $\Po$ from the netlist $\N$ using a rule-based parser.
% Contextual metadata provides concise definitions for domain-specific terms in $\Q$, such as ``negligible ripple,'' to ensure consistent interpretation across simulations.
% }
\noindent \textbf{Planner Agent.} This agent sets up the simulation by (1) generating the initial state $\textsf{S}$ and (2) determining the number of simulations required by the question $\Q$ and forking the state accordingly. To construct $\textsf{S}$, the planner produces two artifacts: contextual metadata for $\Q$ and a base SPICE program $\Po$. Contextual metadata is necessary to handle domain-specific terminology (e.g., ``negligible ripple'' in Fig.~\ref{fig:example}), so the planner retrieves concise definitions using GPT-4o's web search, caches them, and appends them to the question. To construct $\Po$, the planner converts the netlist $\N$ into a SPICE program using a rule based parser. Then, the planner determines the required simulations by identifying distinct circuit conditions in $\Q$, such as comparisons or parameter changes (e.g., the two switch conditions in Fig.~\ref{fig:example} require separate simulations). When runs differ only by a single parameter, SPICE parameter sweeps are used. For each run, the planner creates a scoped sub-question $\Q_i$ specifying a single configuration.\\
\noindent \textit{State Update.} The initial state $\textsf{S}$ is forked into simulation-specific states ${\textsf{S}_{1,0}, \ldots, \textsf{S}_{k,0}}$, each created by replacing $\Q$ with a scoped sub-question $\Q_i$.\\
\noindent\textit{Implementation.} The planner is implemented with a few-shot LLM prompt that generates the required simulations and sub-questions.

% \vspace{-6mm}
\subsection{Simulation Execution}
% \zy{Each simulation state $\textsf{S}_{i,0}$ is processed independently to produce a SPICE program $\Po_i$ that reflects $\Q_i$ and yields valid outputs.
% Since the planner provides only a base program, missing circuit, analysis, and output specifications must be inferred from $\Q_i$ and encoded as edit patches to $\Po$ (Fig.~\ref{fig:patch}).
% The patched program is executed, and failures trigger the repair loop described below.}
After initialization, each simulation state $\textsf{S}_{i,0}$ is processed \textit{independently} to produce a SPICE program $\Po_i$ that reflects $\Q_i$ and yields valid outputs. The planner provides only a base program, so additional specifications must be inferred from the question (Sec.~\ref{sec:challenges}). To this end, $\Po_i$ is constructed through a sequence of patches that encode missing circuit, analysis, and output specifications implied by $\Q_i$ (Fig.~\ref{fig:patch}). These patches modify disjoint sections of the program to avoid interference. The resulting program is then executed; if execution fails, \systemname{} invokes an error-diagnosis agent that iteratively repairs and re-executes the program up to a fixed retry limit.

% \zy{
% \noindent \textbf{Circuit Specification Agent.}
% This agent assigns valid numerical parameters to circuit components.
% It fills parameters by extracting explicit values from $\Q_i$, inferring implicit values from qualitative descriptions using contextual metadata, and adding deterministic defaults when needed.
% }
\noindent \textbf{Circuit Specification Agent.} This agent ensures that all circuit components (e.g., voltage sources, resistors) are assigned valid numerical parameters. While some parameters are extracted from the netlist into the base program $\Po$, many remain unspecified (Sec.~\ref{sec:challenges}). The agent addresses this through three mechanisms:
(1) \textit{Explicit Values.} The agent extracts numerical values stated in the question and maps them to the corresponding parameters (e.g., the “0.2 ms” and “0.3 ms” timing values in Fig.~\ref{fig:example}).
(2) \textit{Implicit Values.} The agent infers unstated values implied by qualitative language in the question, using the contextual enrichment metadata stored in the state (e.g., “negligible ripple” implies sufficiently large capacitance in Fig.~\ref{fig:example}).
(3) \textit{Missing Defaults.} If component parameters are omitted, the agent deterministically adds required defaults (e.g., a default resistor value in Fig.~\ref{fig:example}).

% \zy{
% \noindent \textbf{Analysis Specification Agent.}
% This agent selects the required analysis, including DC, AC, transient analysis, or parameter sweeps, based on the semantics of $\Q_i$.
% }
\noindent \textbf{Analysis Specification Agent.}
This agent selects the appropriate analysis by (1) choosing between DC and AC analysis, (2) determining if transient analysis is required, and (3) configuring parameter sweeps if needed. Analysis selection is guided by domain specific keywords in the question. For example, “steady state” indicates DC analysis, “frequency domain” requires AC analysis, and time varying behavior (Fig.~\ref{fig:example}) triggers transient analysis. When input variation is required, the agent configures parameter sweeps.

% \zy{
% \noindent \textbf{Output Specification Agent.}
% This agent determines which measurements are needed to answer $\Q_i$ and emits the corresponding SPICE directives, including direct measurements, time-specific values, or aggregations.
% }
\noindent \textbf{Output Specification Agent.}
This agent determines which simulation measurements are needed to answer the sub question $\Q_i$ and how to extract them. It decides whether the target quantity (1) can be directly reported, (2) must be measured at a specific time or event, or (3) requires aggregation, and emits the corresponding SPICE directives.\vspace{-3mm}\\\\\
\noindent \textit{State Update.} Each state $\textsf{S}_{i,0}$ is updated with program $\Po_i$ and the complete patch history.

\noindent\textit{Implementation.} All agents are LLMs with role-specific few-shot prompts that generate structured edit patches applied programmatically to $\Po$; invalid patches are discarded. \update{To reduce hallucinations, the circuit specification agent applies self-consistency via 3-sample majority voting.}

% \vspace{-2mm}
\subsubsection{Error Handling.}
% \zy{
% If execution fails, \systemname{} augments the state with simulator error messages and invokes an \textbf{error diagnosis agent}.
% The agent infers the likely root cause, including semantic issues not explicit in the logs, and identifies which specification agents should be re-invoked.
% Only the implicated agents generate corrective patches, avoiding full regeneration and reducing cascading errors.
% This execute--diagnose--repair loop continues until execution succeeds or the retry limit $T$ is reached.
% Successful execution yields final states $\textsf{S}_i^\ast$ containing the final programs $\Po_i^\ast$ and simulation logs.
% }
If execution fails for simulation state $\textsf{S}_{i,0}$, \systemname{} enters an iterative repair loop starting from recovery state $\textsf{S}_{i,1}$. The state is augmented with simulator error messages, and the failed SPICE program is routed to an \textit{error diagnosis agent}. Each repair iteration is indexed by $j$, yielding states $\textsf{S}_{i,j}$.\\
\noindent \textbf{Error Diagnosis Agent.}
This agent infers root causes, including issues \textit{not} explicit in the logs (e.g., "timestep too small" may indicate inconsistent parameters), and
\textit{selectively} re-invokes only the faulty specification agents. This improves efficiency: all agents are invoked during initial construction ($j=0$), while later iterations ($j \geq 1$) invoke only implicated agents (Fig.~\ref{fig:error-workflow-and-patch}). The re-invoked agents receive the error context, current program $\Po_{i,j}$, and edit history, and generate corrective patches yielding an updated state $\textsf{S}_{i,j+1}$ after re-execution. The execute-diagnose-repair loop continues until success or retry limit $T$ is reached.

\noindent \textit{State Update.} Each state $\textsf{S}_{i,j}$ maintains all records of programs, patches, and simulation logs. Successful execution of all simulations yields final state $\textsf{S}_i^*$, containing final program $\Po_i^*$ and its output logs.

\noindent\textit{Implementation.}
This agent is implemented as a bounded-loop with structured diagnostic outputs. %Corrective patches are then generated, applied and the program is re-executed with early termination upon success.

% \vspace{-3mm}
\subsection{Simulation Analysis}
% \zy{
% After all simulations execute successfully, the \textbf{answer generation agent} aggregates and interprets the outputs from $\{\textsf{S}_i^\ast\}_{i=1}^k$ to answer the original question $\Q$.
% It jointly considers the diagram, original question, scoped sub-questions, final SPICE programs, and simulation logs to identify relevant measurements and combine results across simulations.
% The agent uses ReAct prompting with domain-specific computation tools, ensuring that numerical operations are executed symbolically rather than by free-form generation.
% }
\vspace{-1mm}
After all SPICE simulations execute successfully, \systemname{} enters the simulation analysis stage, where the \textit{answer generation agent} aggregates and interprets outputs from the scoped simulations to produce a final answer to the original question $\Q$.

\noindent \textbf{Answer Generation Agent.}  
The agent uses a VLM to jointly reason over (1) the diagram, (2) the original question $\Q$, (3) the scoped sub-questions $\{\Q_i\}$, and (4) for each successful state $\textsf{S}_i^*$, the corresponding SPICE program $\Po_i^*$ and simulation logs. Using these inputs, the agent identifies relevant measurements, interprets them in the context of $\Q$, and combines results across the $k$ simulations to produce the final answer.

\noindent\textit{Implementation.}  
The agent uses ReAct prompting~\cite{yao2022react}, restricting the VLM to a small set of domain-specific computation tools (e.g., Ohm’s law, basic complex arithmetic), which execute all numerical operations symbolically.

\vspace{-2mm}
\section{Evaluation Setup}\label{sec:evaluation-setup}
\vspace{-2mm}
\textbf{Dataset.} Each data point in our setting consists of (1) a circuit diagram, (2) its netlist, and (3) a question requiring mathematical analysis. To the best of our knowledge, \textit{no} existing dataset satisfies all three requirements. We therefore construct two new datasets by augmenting (1) EEE-Bench~\cite{li2025eee} and (2) AMSNet~\cite{shi2025amsnet}.

\noindent \textbf{\eeebench{}}. We adapt EEE-Bench by filtering to electrical circuit questions only, yielding 1,205 data points, and by augmenting each diagram with a corresponding netlist, which is not provided in the original benchmark (details in App.~\ref{app:annotation_errors}).

\noindent \textbf{\amsnet{}}. AMSNet provides paired circuit diagrams and netlists but lacks QA annotations. To avoid costly manual labeling, we automatically construct QA pairs, ensuring answer correctness and non-trivial reasoning via two steps:  
(1) \textit{Base Question Generation.} We generate and execute a SPICE program from the ground-truth netlist, then prompt an LLM to form questions from reported measurements (e.g., ``What is the current through resistor R?''). Since questions are derived from simulation outputs, answers are guaranteed to be correct.  
(2) \textit{Reasoning Augmentation.} We add clauses that introduce reasoning requirements, such as modifying component values or analysis settings (e.g., ``If resistor $R=20k\Omega$, what is the new current?''). Each clause corresponds to a SPICE program edit that is re-executed to produce an updated log from which the answer is derived. This yields 495 QA pairs in the \amsnet{} dataset. To avoid contamination, the LLM used for question generation is distinct from all evaluation baselines (see  App.~\ref{app:annotation_amsnet}). \update{Since \amsnet{} uses SPICE to produce answers, it is structurally aligned with simulation based reasoning, whereas \eeebench{} contains textbook style questions requiring broader reasoning. Still, \amsnet{} preserves the core challenge since \systemname{} must still interpret modification clauses and edit the netlist.} %Consistent with this, \systemname{} does not perform better on \amsnet{} than on \eeebench{} (Sec.~\ref{sec:eval}).}

\noindent \textbf{Simulation Engine.} We use
\texttt{ngspice} (a widely used SPICE simulator) throughout all experiments.\\
\noindent \textbf{Baselines.} To the best of our knowledge, no existing system is tailored for circuit QA (Sec.~\ref{sec:related work}). We therefore evaluate five leading VLMs: \textsc{GPT-5.1}, \textsc{GPT-4o}, \textsc{Claude Sonnet~4}, \textsc{GLM-4.5V} and \textsc{Qwen3-VL-32B-Instruct} under chain-of-thought prompting (\textsc{CoT}), \update{along with a code generation baseline (\textsc{Code}) where the model produces and executes \texttt{SymPy} code to derive and solve circuit equations from the question, netlist and diagram (Prompts in App.~\ref{app:prompts})}. We also compare against prior general-purpose visual QA methods which do not require fine-tuning: \textsc{GIFoMR}~\cite{wang2025glfomr}, \textsc{NoteMR}~\cite{notemr}, \textsc{Reflectiva}~\cite{reflectiva}, and the tool-use reasoning method \textsc{MathSensei}~\cite{das2024mathsensei} (See App.~\ref{app:experiments}).\\
\noindent \textbf{Metric.} We report accuracy. \update{Numeric predictions are correct within tolerance $\epsilon=0.5$}, and textual predictions use a 90\% cosine-similarity threshold. Runtime and token costs are reported in App.~\ref{app:cost}.
% (\texttt{all-MiniLM-L6-v2}).

% \begin{itemize}
%     \item E1: Accuracy
%     \item E2: Design choices
%     \item E3: Error analysis
% \end{itemize}

% Preamble:
% \usepackage{booktabs}

%Our task requires a dataset that satisfies two criteria: (1) circuit diagrams as visual input, and (2) questions that require nontrivial mathematical analysis of the underlying circuit behavior. Existing circuit question answering datasets such as ElectroVizQA and CircuitVQA \cite{ElectroVizQA, CircuitVQA} do not meet the second criterion, as their questions primarily target visual inspection or component counting rather than circuit analysis. We therefore build on two datasets (1) the EEE-Bench dataset \cite{li2025eee}, and (2) the AMSNet dataset \cite{shi2025amsnet}.
\vspace{-2mm}
\section{Experiments and Analysis}\label{sec:eval}
\vspace{-2mm}
We evaluate \systemname{} via the following questions:
\vspace{-2mm}
\begin{packeditemize}
\item \textbf{Q1.} How accurately can \systemname{} answer mathematical questions about electrical circuits?
\item \textbf{Q2.}  What is the impact of each design choice?
\item \textbf{Q3.} What are the main error patterns  of \systemname{}?
\end{packeditemize}
\begin{table}[t]
\centering
\small
\setlength{\tabcolsep}{5pt}

\begin{tabular}{@{}lccc@{}}
\toprule
\multicolumn{4}{c}{\textbf{\eeebench{} Acc. (\%)}} \\
\cmidrule(lr){1-4}
\textbf{Model} & \textbf{\textsc{CoT}} & \textbf{\textsc{Code}} & \textbf{\systemname{}} \\
\midrule
\textsc{GPT 5.1}    & 46.221 & 45.500 & \textbf{83.058} \,($\uparrow$ 36.837) \\
\textsc{Claude S4}  & 51.659 & 31.500 & \underline{81.991} \,($\uparrow$ 30.332) \\
\textsc{GPT 4o}     & 43.151 & 14.030 & 80.829 \,($\uparrow$ 37.678) \\
\textsc{GLM 4.5V}   & 31.662 & 12.540 & 72.116 \,($\uparrow$ \textbf{40.454}) \\
\textsc{Qwen3 I.}   & 41.667 & 27.012 & 80.581 \,($\uparrow$ \underline{38.914}) \\
\midrule
\end{tabular}

\begin{tabular}{@{}lccc@{}}
\multicolumn{4}{c}{\textbf{\amsnet{} Acc. (\%)}} \\
\midrule
\textbf{Model} & \textbf{\textsc{CoT}} & \textbf{\textsc{Code}} & \textbf{\systemname{}} \\
\midrule
\textsc{GPT 5.1}    & 55.791 & 8.000 & \textbf{81.381} ($\uparrow$ 25.590) \\
\textsc{Claude S4}  & 57.205 & 8.791 & \underline{80.658} ($\uparrow$ 23.453) \\
\textsc{GPT 4o}     & 29.899 & 8.951 & 79.053 ($\uparrow$ \underline{49.154}) \\
\textsc{GLM 4.5V}   & 46.296 & 2.198 & 78.407 ($\uparrow$ 32.111) \\
\textsc{Qwen3 I.}   & 21.953 & 3.297 & 80.606 ($\uparrow$ \textbf{58.653}) \\
\bottomrule
\end{tabular}

\vspace{-2mm}
\caption{Baselines vs. \systemname{} accuracy (\%); $\uparrow$ absolute improvement over \textsc{CoT}, best score and largest gains in \textbf{bold}, second best \underline{underlined}.}
\label{tab:best_accuracy}
\end{table}

\begin{table}[t]
\centering
\tiny
\setlength{\tabcolsep}{2pt}
\renewcommand{\arraystretch}{1.15}

\resizebox{\linewidth}{!}{%
\begin{tabular}{@{}lccccc@{}}
\toprule
\textbf{Dataset}
& \textsc{NoteMR}
& \textsc{Reflec.}
& \textsc{Math.}
& \textsc{GIFoMR}
& \textsc{\systemname{}} \\
\midrule
\textbf{\eeebench{}}
& \begin{tabular}[c]{@{}c@{}}38.116\\{\tiny($\downarrow$44.9)}\end{tabular}
& \begin{tabular}[c]{@{}c@{}}22.255\\{\tiny($\mathbf{\downarrow}$\textbf{60.8})}\end{tabular}
& \begin{tabular}[c]{@{}c@{}}54.378\\{\tiny($\downarrow$28.7)}\end{tabular}
& \begin{tabular}[c]{@{}c@{}}36.633\\{\tiny($\downarrow$46.4)}\end{tabular}
& \begin{tabular}[c]{@{}c@{}}83.058\\{\tiny(--)}\end{tabular} \\

\textbf{\amsnet{}}
& \begin{tabular}[c]{@{}c@{}}46.670\\{\tiny($\downarrow$34.7)}\end{tabular}
& \begin{tabular}[c]{@{}c@{}}16.969\\{\tiny($\mathbf{\downarrow}$\textbf{64.4})}\end{tabular}
& \begin{tabular}[c]{@{}c@{}}58.989\\{\tiny($\downarrow$22.4)}\end{tabular}
& \begin{tabular}[c]{@{}c@{}}34.725\\{\tiny($\downarrow$46.7)}\end{tabular}
& \begin{tabular}[c]{@{}c@{}}81.381\\{\tiny(--)}\end{tabular} \\
\bottomrule
\end{tabular}%
}
\vspace{-2mm}
\caption{\systemname{} vs prior work accuracy (\%); $\downarrow$ is the gap to \systemname{} (largest gap in \textbf{bold}).}\label{tab:prior_work_comparison}
\end{table}

\vspace{-3.8mm}
\subsection{Q1. Comparative Performance Analysis.}
Table~\ref{tab:best_accuracy} reports accuracy on \eeebench{} and \amsnet{} for zero-shot \textsc{CoT} and \textsc{Code} baselines, while Table~\ref{tab:prior_work_comparison} compares \systemname{} against prior work. \textbf{First, \systemname{} consistently outperforms all baselines.} It achieves the highest accuracy on both datasets, with 83.1\% on \eeebench{} and 81.4\% on \amsnet{}. Relative to zero-shot \textsc{CoT} baselines, \systemname{} improves accuracy by 30.3--40.5 points on \eeebench{} and 23.5--58.7 points on \amsnet{}. Compared to prior work, the gains range from 28.7--60.8 points on \eeebench{} and 22.4--64.4 points on \amsnet{}, indicating that the improvements stem from \systemname{}'s design.
\textbf{Second, weaker base models benefit the most.} Models with lower \textsc{CoT} performance see the largest gains--up to 40.5 points on \eeebench{} and 58.6 points on \amsnet{}. On \eeebench{}, these gains are most pronounced for open-source models \textsc{GLM 4.5V} and \textsc{Qwen3}, indicating that \systemname{} compensates for the models' limited reasoning ability through its structured design. On \amsnet{}, the worst-performing baselines (\textsc{GPT-4o} and \textsc{Qwen}) often failed to produce an answer, instead emitting placeholder tokens; \systemname{} avoids this failure mode by increasing confidence in predictions by using simulations. \textbf{Third, gains persist even for the strongest baselines.} \textsc{GPT 5.1} and \textsc{Claude} still improve by $>$23 points on both datasets. \systemname~also outperforms the best prior work, \textsc{MathSensei}, by $>$22 points.
\update{\textsc{Code} substantially underperforms (\systemname{} gains of up to +66.8 on \eeebench{} and +76.2 on \amsnet{}.)}
\vspace{-3mm}
\subsection{Q2. Design Choice Validation.}
\label{sec:eval:ablations}
\vspace{-1mm}
We conduct ablations to assess \systemname{}’s design:\\
\noindent \textbf{(1) Input.} Answering circuit questions involves (i) extracting netlist from diagrams and (ii) synthesizing a SPICE program from the question. \update{We evaluate the fully end-to-end setting using automatically extracted netlists instead of gold annotations, directly measuring the impact of VLM-based netlist extraction; an orthogonal task to \systemname{}. We also compare three input settings to evaluate the effect of providing netlists to baselines. For \systemname{}, we measure the effect of including the diagram in the input (Table~\ref{tab:ablation1}).} For baselines, we observe that
\textbf{using netlists as input does not consistently improve accuracy.} On \eeebench{}, netlists reduce accuracy by 0.3--5.5 points for all but \textsc{GPT} models. On \amsnet{}, netlists yield gains of 11.3--18.1 points when comparing \N+\Q{} and \N+\Q+\D{} to \Q+\D{}. 
\begin{table}[t]
\centering
\small
\setlength{\tabcolsep}{3pt}
\resizebox{\linewidth}{!}{%
\begin{tabular}{@{}lccccc@{}}
\toprule
& \multicolumn{3}{c}{\textsc{CoT}} & \multicolumn{2}{c}{\textbf{\systemname{}}} \\
\cmidrule(lr){2-4}\cmidrule(lr){5-6}
\textbf{Model} & \N+\Q & \N+\Q+\D & \Q+\D & \N+\Q & \N+\Q+\D \\
\midrule
\multicolumn{6}{c}{\textbf{\eeebench{} Acc. (\%)}} \\
\midrule
\textsc{GPT 5.1}   & 45.960 & \textbf{46.221} & 44.657 & 78.226 & 83.058 \\
\textsc{Claude S4} & 46.130 & 47.077 & \textbf{51.659} & 71.429 & 81.991 \\
\textsc{GPT 4o}    & 35.103 & \textbf{43.151} & 40.240 & 76.923 & 80.829 \\
\textsc{GLM 4.5V}  & 26.424 & 28.587 & \textbf{31.662} & 63.147 & 72.116 \\
\textsc{Qwen3 I.}  & 36.645 & 40.384 & \textbf{41.667} & 66.667 & 80.581 \\
\midrule
% \multicolumn{6}{l}{\textit{auto-extracted \N{}:}} \\
\textsc{GPT 5.1}(auto-extracted \N{})$^\dagger$ & {--} & 44.071 & \multicolumn{2}{c}{--} & 65.283 \\
\midrule
\multicolumn{6}{c}{\textbf{\amsnet{} Acc. (\%)}} \\
\midrule
\textsc{GPT 5.1}   & \textbf{55.791} & 52.492 & 41.178 & 67.857 & 81.381 \\
\textsc{Claude S4} & \textbf{57.205} & 54.781 & 39.125 & 67.742 & 80.658 \\
\textsc{GPT 4o}    & 24.242 & 25.758 & \textbf{29.899} & 75.789 & 79.053 \\
\textsc{GLM 4.5V}  & \textbf{46.296} & 45.286 & 29.529 & 64.118 & 78.407 \\
\textsc{Qwen3 I.}  & \textbf{21.953} & 20.303 & 21.717 & 69.692 & 80.606 \\
\midrule
\textsc{GPT 5.1}(auto-extracted \N{})$^\dagger$ & {--} & 30.516 & \multicolumn{2}{c}{--} & 50.701 \\
\bottomrule
\end{tabular}%
}
\vspace{-5pt}
\caption{Accuracy across input settings (\%); \Q+\D{} does not apply to \systemname{}; Best baseline in \textbf{bold}. $^\dagger$measured for best performing model \textsc{GPT 5.1} for \systemname{}.}
\label{tab:ablation1}
\end{table}
%  \textsc{Qwen3 I.}=\textsc{Qwen3-VL-32B-Instruct};\textsc{Claude S4}=\textsc{Claude Sonnet 4}.
\begin{table}[t]
\centering
\small
\resizebox{0.99\linewidth}{!}{
\setlength{\tabcolsep}{4pt}

\begin{tabular}{lcccc}
\toprule
\textbf{Model} 
& \multicolumn{2}{c}{\textbf{\eeebench{}}} 
& \multicolumn{2}{c}{\textbf{\amsnet{}}} \\
\cmidrule(lr){2-3}\cmidrule(lr){4-5}
& \textbf{Direct} & \textbf{\systemname{}} 
& \textbf{Direct} & \textbf{\systemname{}} \\
\midrule
\textsc{GPT 5.1}               
& 26.04 & \textbf{97.93} (×3.76) 
& 56.36 & \textbf{96.57} (×1.71) \\

\textsc{GPT 4o}                
& 13.61 & 96.43 (×7.09) 
& 41.81 & 95.96 (×2.29) \\

\textsc{Claude S4}       
& 16.70 & 96.82 (×5.80) 
& 40.20 & 98.18 (×2.44) \\

\textsc{GLM 4.5V}              
& 0.07  & 86.72 (×\textbf{1239}) 
& 13.24 & 96.36 (×\textbf{7.27}) \\

\textsc{Qwen3 I.} 
& 5.00  & 92.03 (×18.41) 
& 41.21 & 95.35 (×2.31) \\
\bottomrule
\end{tabular}

}
\vspace{-2mm}
\caption{SPICE program executability (\%) under direct prompting and \systemname{}; × multiplicative improvement over direct prompting; best or largest gains in \textbf{bold}.}
\label{tab:spice_executability}
\end{table}
\update{For \systemname{}, including the diagram  improves accuracy on both datasets. Using automatically extracted instead of gold netlists reduces accuracy by 17.8 points on \eeebench{} and 30.7 points on \amsnet{}, showing that netlist extraction quality remains an important bottleneck. Nevertheless, under the same setting, \systemname{} still outperforms the \textsc{CoT} baseline by 21.2 points on \eeebench{} (65.3\% vs.\ 44.1\%) and 20.2 points on \amsnet{} (50.7\% vs.\ 30.5\%), confirming that its pipeline advantages persist. Discussion of netlist extraction accuracy is in App.~\ref{app:model_progression}.}
% \textbf{Second, the utility of netlists is dataset dependent.} On \eeebench{}, netlists have little impact on accuracy, whereas on \amsnet{}, performance improves when netlists are provided. This follows from dataset design: \eeebench{} questions are drawn from textbooks and often require theoretical reasoning beyond the circuit itself, while \amsnet{} questions are derived from simulation outputs and are more tightly focused on the circuit. %and its dense diagram annotations further limit diagram utility.
% \arc{the above point is unclear - diagram and netlist are the equivalent}

\noindent \textbf{(2) Multi-stage Pipeline.}
Table~\ref{tab:spice_executability} shows that direct prompting, which generates a SPICE program in a single step, frequently produces non-executable programs (executability as low as 0.07\%), whereas \systemname{} achieves over 95\% executability.\\
\noindent \update{\textbf{(3) Agent Ablation.}
Removing any single specialized agent (or web search/self-consistency in circuit specification agent) consistently degrades accuracy on both datasets (Table~\ref{tab:ablation_agents}, App.~\ref{app:ablation}).}

\vspace{-1mm}
\subsection{Q3. Error Analysis for Reliability.}
% \vspace{2mm}
\label{sec:error-analysis}

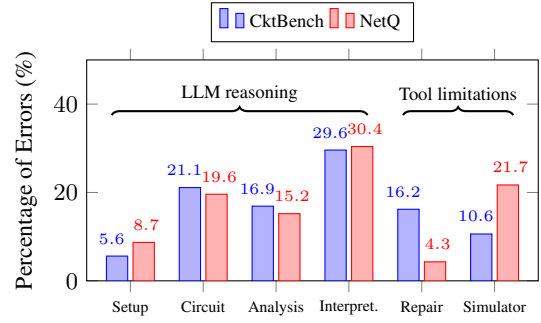
\begin{figure}[t]
\centering
\small
\begin{tikzpicture}
\begin{axis}[
    ybar,
    bar width=8pt,
    width=0.98\linewidth,
    height=4.5cm, % reduced
    ymin=0,
    ymax=50,
    ylabel={Percentage of Errors (\%)},
    symbolic x coords={Setup,Param,Analysis,Interp,Repair,Simulator},
    xtick=data,
    xticklabels={Setup,Circuit,Analysis,Interpret.,Repair,Simulator},
    xticklabel style={rotate=0, anchor=north, align=center, font=\tiny}, % non angular
    % ymajorgrids=true,
    % grid style=dashed,
    enlarge x limits=0.12,
    legend style={at={(0.5,1.08)}, anchor=south, legend columns=2, font=\scriptsize}, % closer
]

% \eeebench{} (shift labels slightly left)
\addplot[
    blue, fill=blue!30,
    nodes near coords,
    point meta=y,
    every node near coord/.append style={
        font=\tiny,
        anchor=south,
        yshift=1.5pt,
        xshift=-2pt,
        /pgf/number format/fixed,
        /pgf/number format/precision=1
    }
] coordinates {
    (Setup,5.6)
    (Param,21.1)
    (Analysis,16.9)
    (Interp,29.6)
    (Repair,16.2)
    (Simulator,10.6)
};

% \amsnet{} (shift labels slightly right)
\addplot[
    red, fill=red!30,
    nodes near coords,
    point meta=y,
    every node near coord/.append style={
        font=\tiny,
        anchor=south,
        yshift=1.5pt,
        xshift=1pt,
        /pgf/number format/fixed,
        /pgf/number format/precision=1
    }
] coordinates {
    (Setup,8.7)
    (Param,19.6)
    (Analysis,15.2)
    (Interp,30.4)
    (Repair,4.3)
    (Simulator,21.7)
};

\legend{\eeebench, \amsnet}
% half of the total group width (tune if you change bar width/shifts)
\newcommand{\halfgroupshift}{7pt}

% place braces closer to the bars
\draw[decorate, decoration={brace, amplitude=3pt}, thick]
  ([xshift=-\halfgroupshift]axis cs:Setup,37.5) --
  ([xshift= \halfgroupshift]axis cs:Interp,37.5)
  node[midway, yshift=8pt, font=\scriptsize]{LLM reasoning};

\draw[decorate, decoration={brace, amplitude=3pt}, thick]
  ([xshift=-\halfgroupshift]axis cs:Repair,37.5) --
  ([xshift= \halfgroupshift]axis cs:Simulator,37.5)
  node[midway, yshift=8pt, font=\scriptsize]{Tool limitations};

\end{axis}
\end{tikzpicture}
\vspace{-0.3cm}
\caption{Distribution of primary failure modes among incorrect predictions. Examples are in App.~\ref{app:RQ3}.}
\label{fig:error_breakdown}
\end{figure}

\systemname{} improves reliability in two ways. First, correct answers are \textit{verifiable}: all outputs are grounded in SPICE simulations. Second, by decomposing the task into explicit stages, failures can be traced to their root causes. We assign each incorrect prediction to a single \emph{failure mode} (Fig.~\ref{fig:error_breakdown}). 
%  \textcolor{blue!60}{Blue} bars correspond to \eeebench{}, \textcolor{red!60}{red} bars to \amsnet{}.

\noindent \textbf{LLM Reasoning Errors.}
These errors arise from incorrect reasoning despite correct simulation execution, including: \emph{simulation setup errors}, where the planner selects the wrong number of simulations (5.6\% on \eeebench{}, 8.7\% on \amsnet{}); \emph{circuit specification errors}, caused by incorrect parameter edits (21.1\%,19.6\%); \emph{analysis specification errors}, from wrong simulation analyses (16.9\%,15.2\%); and \emph{result interpretation errors}, where correct outputs are misinterpreted by the answer generation agent (29.4\%,30.4\%).\\
\noindent \textbf{Tool Limitation Errors.}
These errors stem from limitations of the LLMs and simulator. \emph{Execution correction errors} occur during iterative repair when correct edits are overwritten (16.2\% on \eeebench{}, 4.3\% on \amsnet{}). \emph{Simulator errors} arise from limitations of \texttt{ngspice}, whose simplified device models can deviate from ideal behavior (10.6\%, 21.7\%).

% \noindent \textbf{Discussion.}
% Three trends emerge. First, over 75\% of failures stem from reasoning errors, making reasoning the primary bottleneck. Second, many failures reflect limited domain understanding despite correct simulation execution. Third, error patterns differ across datasets: \eeebench{} exhibits more execution correction errors, while \amsnet{} shows more simulator related failures. Importantly, \systemname{}’s staged architecture enables these errors to be isolated and analyzed. %These findings highlight stronger domain-aware reasoning and state consistency as key directions for future work.

\vspace{-0.2cm}
\section{Related Work}\label{sec:related work}
\vspace{-2mm}
\noindent\textbf{Electrical Circuit Analysis.}
To the best of our knowledge, \textit{no} prior work addresses mathematically grounded QA over electrical circuit diagrams. Existing efforts focus on vision and extraction~\cite{CircuitVQA, ElectroVizQA, shi2025amsnet}, benchmarks~\cite{li2025eee}, or use SPICE for unrelated tasks~\cite{nau2025spiceassistant}.\\
\noindent\textbf{General Diagram Question Answering.}
Prior diagram QA either emphasizes semantic interpretation~\cite{wang2024cog, wang2025glfomr, wang2024charxiv, farahani2025chart, notemr, reflectiva} or tool-augmented reasoning~\cite{bauer2023neuro, suri2025follow, plotqa, das2024mathsensei, toolvqa}. Unlike prior methods that explicitly formulate equations in code, \systemname{} synthesizes simulator-native SPICE programs and relies on a physics-based engine for equation selection and solving. See App.~\ref{app:related work} for more details.

\vspace{-2mm}
\section{Conclusion}
\vspace{-2mm}
We have presented \systemname{}, a system that answers electrical circuit questions by grounding reasoning in executable, physics-based simulations. Its execution-guided design improves both accuracy and reliability over LLM baselines, demonstrating the effectiveness of simulator-native reasoning for complex circuit QA.

\section*{Limitations}
Our work focuses on electrical circuit problems, where SPICE-based simulation provides a natural and effective tool for grounding quantitative reasoning. Many other problem domains in electrical and electronic engineering involve different types of analysis tools and formalisms, which we leave to future work.
\section*{Ethical considerations}
All datasets used in this work are publicly available and released under open licenses. The tools and models employed are authorized for research purposes and have been used in accordance with their intended terms. Detailed license information is provided in Appendix~\ref{appendix:artifact}. All experiments were performed strictly for research and evaluation. To the best of the authors’ knowledge, this research does not introduce any ethical risks.

\bibliography{custom}

\appendix
\clearpage
\section{Experiments and Analysis (Cntd.)}\label{app:experiments}
\begin{figure*}[t]
    \centering
    \includegraphics[width=1\linewidth]{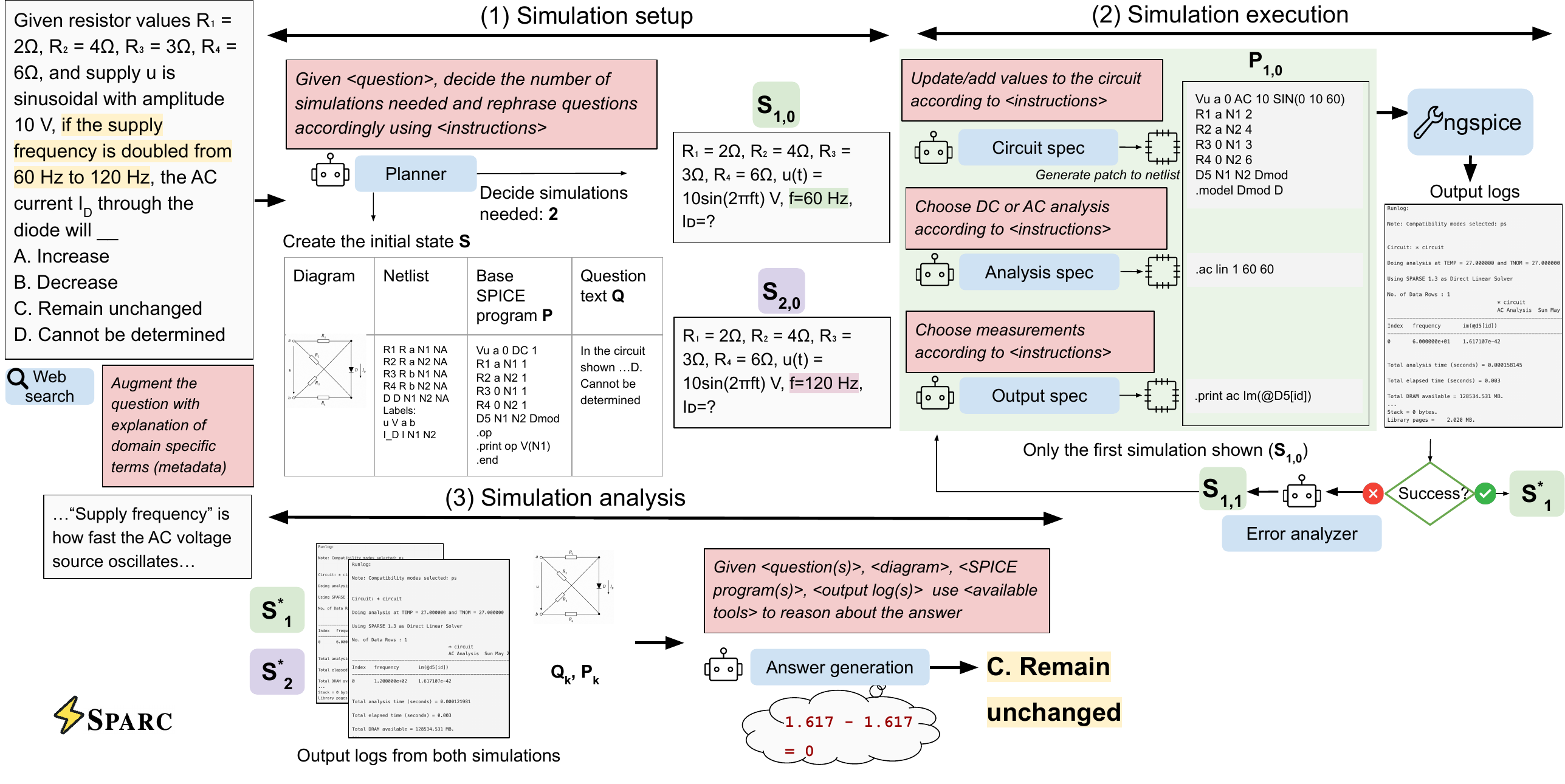}
    \caption{\update{End to end example of \systemname{}. Given a question, it is first augmented with explanations of domain specific terms using web search results (e.g., "supply frequency"). The \textit{planner} then creates the initial state \textsf{S} containing the diagram \D{}, netlist \N{}, and question \Q{}, constructs the base program \Po{}, and determines that two simulations are required, producing states \( \textsf{S}_{1,0} \) and \( \textsf{S}_{2,0} \). For simplicity, only the execution flow for \( \textsf{S}_{1,0} \) is shown. The \textit{circuit}, \textit{analysis}, and \textit{output} specification agents independently generate patches for disjoint sections of \Po{}, producing \( \Po_{1,0} \), which is executed with \texttt{ngspice}. If execution succeeds, the state is updated to the final state \( \textsf{S}_1^* \). Otherwise, an \textit{error analyzer} identifies the faulty section and selectively re-invokes the responsible agent for repair, yielding updated states such as \( \textsf{S}_{1,1} \). This repeats until execution succeeds or retry limit \(T\) is reached. Finally, the \textit{answer generation} agent aggregates the successful simulation logs and uses tool based calculations to produce the final answer.}}
    \label{fig:end-to-end-example}
\end{figure*}
\noindent \textbf{Experimental Setting.} We report model sizes, computational budget, and infrastructure details for all experiments. The evaluated models include \textsc{GPT-5.1}, \textsc{Claude Sonnet 4}, \textsc{GPT-4o} (parameter counts not publicly disclosed), \textsc{Qwen3-VL-32B Instruct} (32B parameters), and \textsc{GLM-4.5V} (108B parameters). All development and evaluation runs were conducted via the OpenRouter API. Local compute was used only for orchestration and logging, using 48 core Intel Xeon Silver 4310 CPUs with 128 GB RAM on Ubuntu 24.04.2 LTS. These details contextualize the computational scale of the experiments.

\noindent \textbf{Baselines.} Since the question-answering components of \textsc{GIFoMR} \cite{wang2025glfomr}, \textsc{NoteMR} \cite{notemr} and \textsc{MathSensei} \cite{das2024mathsensei} are model-agnostic, we instantiate them using our strongest performing model for a fair comparison. For \textsc{NoteMR}, we follow the original pipeline by using LLaVA for visual grounding and Grad-CAM based region selection, and then use our best-performing vision language model, \textsc{GPT-5.1} for the final QA stage. For \textsc{MathSensei}, we compare against its best-performing configuration (PG+SG) while replacing the base language model with \textsc{GPT-5.1}. For \textsc{GIFoMR}, we implement the four-stage pipeline described in the paper using the provided prompts, with \textsc{GPT-5.1} as the underlying model. Together, these constitute the strongest and most relevant existing baselines for our task.

We also construct chain of thought prompting baselines for all models, along with a code generation baseline where the model produces \texttt{SymPy} \cite{sympy} code to derive and solve the governing circuit equations from the netlist and circuit diagram, and then executes the generated code to obtain the final answer. We use \texttt{SymPy} because circuit QA fundamentally requires symbolic mathematical reasoning over equations derived from physical laws such as Ohm's law and Kirchhoff's laws, making symbolic equation manipulation and solving a natural fit for the task. 

\subsection{\update{Progress on Automated Netlist Extraction Accuracy}}
\label{app:model_progression}

\begin{figure}[htbp]
  \centering
  \includegraphics[width=1\linewidth]{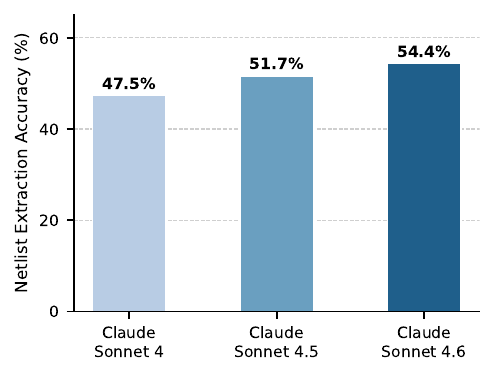}
  \caption{\update{Netlist extraction F1 on \eeebench{} across Claude Sonnet generations. Each bar reports the F1 score for circuit component identification per diagram.}}
  \label{fig:component_recall_progress}
\end{figure}

\update{Prior work such as \cite{shi2025amsnet} reports up to 90 F1 score on netlist extraction. Figure~\ref{fig:component_recall_progress} further shows steady improvements in component recognition performance across successive Claude Sonnet generations on the \eeebench{} benchmark. Component identification F1 improves from 47.5\% (Sonnet~4) to 51.7\% (Sonnet~4.5) and 54.4\% (Sonnet~4.6), suggesting that advances in visual understanding directly improve circuit extraction accuracy. Moreover, our manually annotated \eeebench{} dataset with paired circuit diagrams and netlists, together with existing datasets such as AMSNet2.0~\cite{shi2025amsnet}, provides a foundation for future fine tuning of circuit extraction models.}

\subsection{\update{Runtime and Cost}}
\label{app:cost}
\update{Table~\ref{tab:cost_runtime} reports per-query wall-clock time and total token usage for \textsc{CoT}, \textsc{Code}, and \systemname{}. \systemname{} incurs higher latency and token cost than single-step baselines, reflecting the overhead of its multi-agent pipeline. Latency varies substantially across models: \textsc{GLM-4.5V} is the slowest due to its larger size, while \textsc{GPT-4o} is the fastest \systemname{} configuration. Token usage for \systemname{} is dominated by the answer generation agent, which receives full simulation logs as context.}

% Single-sample wall-clock time and token usage per query.
% Tokens reported as total (input + output) in thousands (K).
\begin{table}[t]
\centering
\small
\setlength{\tabcolsep}{4pt}
\resizebox{\linewidth}{!}{%
\begin{tabular}{@{}lcccccc@{}}
\toprule
& \multicolumn{3}{c}{\textbf{Time (s)}}
& \multicolumn{3}{c}{\textbf{Tokens (K)}} \\
\cmidrule(lr){2-4}\cmidrule(lr){5-7}
\textbf{Model}
  & \textsc{CoT} & \textsc{Code} & \systemname{}
  & \textsc{CoT} & \textsc{Code} & \systemname{} \\
\midrule
\textsc{GPT 5.1}   &   1.3 &   9.1 &  67.2 & 0.5 & 1.1 & 27.2 \\
\textsc{Claude S4} &   9.6 &  18.1 &  55.7 & 0.9 & 2.1 & 47.5 \\
\textsc{GPT 4o}    &   4.8 &   7.7 &  20.2 & 0.9 & 1.2 & 26.5 \\
\textsc{GLM 4.5V}  &   1.3 &  14.0 & 208.3 & 0.5 & 1.0 & 32.6 \\
\textsc{Qwen3 I.}  &   6.7 &   8.9 &  47.7 & 0.9 & 1.1 & 26.3 \\
\bottomrule
\end{tabular}%
}
\vspace{-4pt}
\caption{Per-query wall-clock time (seconds) and total token usage (input\,+\,output, in thousands) for \textsc{CoT}, \textsc{Code}, and \systemname{}, shown for per dataset sample.}
\label{tab:cost_runtime}
\end{table}

\subsection{Ablation Study}
\label{app:ablation}

\noindent\textbf{\update{Agent Ablation.}}
\update{Table~\ref{tab:ablation_agents} reports accuracy when each specialized agent is individually removed from the \systemname{} pipeline. We ablate the circuit specification agent, analysis specification agent, output specification agent, and the web-search component of the planner (which supplies domain-context enrichment).}

% Agent ablation: accuracy when each specialized agent is removed from \systemname{}.
\begin{table}[t]
\centering
\small
\setlength{\tabcolsep}{5pt}
\resizebox{\linewidth}{!}{%
\begin{tabular}{@{}lcc@{}}
\toprule
\textbf{Configuration} & \textbf{\eeebench{}} & \textbf{\amsnet{}} \\
\midrule
\systemname{} (full)                        & 83.058   & 81.381   \\
\midrule
w/o Circuit Spec.\ Agent                    &   74.121 &  65.104  \\
w/o Analysis Spec.\ Agent                   & 71.712   & 62.890   \\
w/o Output Spec.\ Agent                     & 70.599 & 61.190   \\
w/o Error Analyzer Agent                     & 73.612 & 68.351   \\
w/o Web Search (Planner)                    & 70.225   & 67.495   \\
w/o No SC in Circuit Spec.\ Agent                    & 78.781   & 76.452   \\
\bottomrule
\end{tabular}%
}
\vspace{-4pt}
\caption{Accuracy (\%) of \systemname{} ablations using \textsc{GPT 5.1}. Each row removes one agent from the full pipeline (except for the last row, which only removes self consistency from the circuit specification agent). We report results only for the best performing model in \systemname{}, \textsc{GPT 5.1}; SC=self-consistency.}
\label{tab:ablation_agents}
\end{table}

\subsection{RQ3. Error Analysis Cntd.}\label{app:RQ3}

We analyze all incorrect predictions and assign each to a single \emph{primary failure mode} aligned with the stages of the \systemname{} pipeline. Figure~\ref{fig:error_breakdown} summarizes the distribution.

To systematically identify recurring patterns, we additionally cluster incorrect predictions using Jina embeddings~\cite{jina} and HDBSCAN~\cite{mcinnes2017hdbscan}, producing 75 semantically coherent clusters. We manually inspect these clusters and consolidate them into the taxonomy below.

\paragraph{Simulation Setup Errors.}
These errors occur when the system incorrectly decomposes a question into simulation tasks. A common case involves optimization over continuous parameters. For example, questions asking which phase angle maximizes a voltage at a switching time require sweeping \( \theta \) over a continuous range. However, the system sometimes treats the task as a finite comparison and evaluates only a few discrete angles. As a result, the correct value, such as \( \theta = -45^\circ \), may never be tested. We observe this pattern in 5.6\% of failures on \eeebench{} and 8.7\% on \amsnet{}.

\paragraph{Parameter Specification Errors.}
These errors occur when the circuit structure and analysis type are correct, but numerical values or constraints are applied incorrectly. Typical failures include updating the wrong component, modifying irrelevant elements, or ignoring stated assumptions, causing the simulated circuit to diverge from the intended specification.

For example, in questions asking for the steady-state operating voltage of a transistor circuit, the system may correctly select steady-state analysis but modify an output resistor simply because it is mentioned in the question, even though it does not affect the queried voltage. This reflects incorrect parameter relevance reasoning rather than an execution failure. We observe this pattern in 21.1\% of failures on \eeebench{} and 19.6\% on \amsnet{}.

\paragraph{Analysis Selection Errors.}
These failures occur when the model selects the wrong simulation analysis despite correct circuit extraction. Common cases include choosing transient analysis when a DC operating point suffices, confusing AC and DC behavior, or selecting analyses based on components irrelevant to the queried quantity. The simulator therefore produces valid outputs for a physically mismatched analysis.

These failures are not caused by simulator or pipeline limitations. \systemname{} exposes the relevant analysis modes and executes them faithfully. The root cause is incomplete electrical engineering knowledge in the LLM, particularly difficulty identifying when certain components or dynamic behaviors are irrelevant under a given operating regime. We observe this pattern in 16.9\% of failures on \eeebench{} and 15.2\% on \amsnet{}.

\paragraph{Result Interpretation Errors.}
This is the largest category, accounting for 29.4\% of failures on \eeebench{} and 30.4\% on \amsnet{}. These errors occur when simulations execute correctly, but the final reasoning over simulator outputs is incorrect. Common issues include polarity mistakes, incorrect analytical formulas, and high-level misinterpretations of otherwise correct outputs.

For example, in one case, the simulation shows that both input and output voltages are positive, indicating that the circuit does not invert the signal. The model nevertheless concludes that the output is phase inverted. The failure therefore lies in interpretation rather than simulation or arithmetic.

\paragraph{Error Correction Failures.}
These errors arise during iterative repair. Although agents receive the full edit history, previously correct edits are sometimes overwritten or reverted, producing circuits that remain executable but semantically incorrect. This issue is more pronounced for models with shorter effective context windows, such as \textsc{GLM-4.5V}, which struggle to preserve earlier edits across repair iterations.

We observe this pattern in 16.2\% of failures on \eeebench{} and 4.3\% on \amsnet{}. The gap likely reflects dataset structure. \amsnet{} questions are generated directly from simulations, making required updates more explicit, whereas \eeebench{} often requires reasoning over implicit constraints, increasing the likelihood of inconsistent repairs.

\paragraph{Simulator Limitations.}
Some failures stem from limitations of the underlying simulator rather than the reasoning pipeline. Circuit simulators such as ngspice rely on simplified device models and numerical solvers, limiting fidelity. Approximate models for idealized or complex components, such as ideal diodes or switches, may not fully match intended theoretical behavior. Small modeling differences can therefore produce outputs that diverge from expected values. We observe this pattern more frequently in \amsnet{} (21.7\%) than in \eeebench{} (10.6\%), consistent with the simulator-centric construction of \amsnet{}.

\paragraph{Discussion.}
The analysis reveals three main trends. First, most failures occur after executable simulations are produced: result interpretation and parameter specification errors together account for over half of all failures, making post-simulation reasoning the primary bottleneck. Second, many errors reflect limited domain understanding rather than system design flaws: \systemname{} exposes the necessary tools and executes them faithfully, while the model struggles to identify relevant parameters, analyses, and behaviors. Third, error distributions differ across datasets: \eeebench{} exhibits more correction failures due to implicit assumptions and multi-step reasoning, whereas \amsnet{} exhibits more simulator-related failures due to its simulator-centric construction. These findings suggest future work should focus on domain-aware reasoning and stronger state consistency across repair iterations.
\section{SPICE syntax background}
\label{app:SPICE}
First, ngspice programs describe the \textbf{structure of the circuit}, including its components, connections, and fixed values. For example, the following statements define a DC voltage source and a resistor connected between two nodes:
\begin{verbatim}
V1 a 0 DC 10
R1 a b 5
\end{verbatim}

Second, ngspice allows the behavior of \textbf{nonlinear or complex components} to be defined explicitly using device models. These models specify how components such as diodes, switches, or transistors behave under different operating conditions and are referenced by individual circuit elements:
\begin{verbatim}
.model DIO D(Is=1e-12 N=1)
D1 a b DIO
\end{verbatim}

Third, an ngspice program must specify \textbf{how the circuit should be analyzed}. Analysis primitives determine whether the simulator computes a steady state operating point, frequency response, or time varying behavior:
\begin{verbatim}
.op
.ac lin 1 {f} {f}
.tran 1u 10m
\end{verbatim}

In addition, ngspice provides primitives for \textbf{controlling and varying simulations}. These include defining symbolic parameters, setting initial conditions, and sweeping parameters or sources to explore multiple operating scenarios:
\begin{verbatim}
.param Rload=10
.ic V(out)=5
.dc Rload 5 20 5
\end{verbatim}

Finally, ngspice programs specify \textbf{what outputs to measure}. Measurement primitives extract quantities of interest from simulation results, such as voltages, currents, or aggregated statistics over time:
\begin{verbatim}
.print tran V(out)
.meas tran Vavg AVG V(out) FROM=1m TO=5m
\end{verbatim}

\section{\systemname{} Examples}
This section presents end-to-end canonical examples illustrating how \systemname{} operates on real world problems. Figure~\ref{fig:end-to-end-example} shows one such example.

\subsection{Agent-Level Examples}
\label{app:agent-examples}

\tcbset{sparcex/.style={%
  fonttitle=\sffamily\scriptsize\bfseries,colback=gray!8,colframe=gray!50,%
  left=3pt,right=3pt,top=1pt,bottom=1pt,boxrule=0.4pt,arc=2pt,%
  before skip=2pt,after skip=3pt,breakable}}

The following examples illustrate the behaviour of each \systemname{} agent across different question types. For agents that modify the base program~$\Po$, boxes labelled \textit{Base (excerpt)} show the relevant portion of~$\Po$ before the agent acts, and boxes labelled \textit{Patch} show the edits the agent produces. The final program is obtained by applying the patch to~$\Po$.

%----------------------------------------------------------------------
\subsubsection{Planner Agent}
%----------------------------------------------------------------------
The planner reads the question and netlist, decides how many independent simulations are required, and writes a focused sub-question for each run. Crucially, it does not touch the SPICE program itself, that is left to the specification agents.

\paragraph{P1: two-run decomposition for a parameter-change scenario.}
\textbf{Q:} In the circuit shown in figure, it is known that \( R_1 = 2\Omega \), \( R_2 = 4\Omega \), \( R_3 = 3\Omega \), \( R_4 = 6\Omega \), and \( R_5 = 10\Omega \). If \( R_5 \) is increased to twice its original value, \( I_5 \) will ( )

This question requires comparing \(I_5\) under two resistance settings. The planner decomposes the task into two runs: one for the original circuit and one for the modified circuit.

\begin{tcolorbox}[sparcex,title={Planner output}]
{\ttfamily\scriptsize
num\_runs 2\\
run 1: $R1 = 2\Omega, R2 = 4\Omega,$\\
\phantom{run 1: }$R3 = 3\Omega, R4 = 6\Omega$, and $R5 = 10\Omega.$\\
\phantom{run 1: }What is I5?\\
run 2: $R1 = 2\Omega, R2 = 4\Omega$,\\
\phantom{run 2: }$R3 = 3\Omega, R4 = 6\Omega$, and $R5 = 20\Omega.$\\
\phantom{run 2: }What is I5?}
\end{tcolorbox}

The answer generation agent later compares the two computed values of \(I_5\) to determine how the current changes when \(R_5\) is doubled.

\paragraph{P2: single-run for a DC equivalent-circuit query.}
\textbf{Q:} Find the Norton equivalent resistance $R_N$ seen at terminals a--b.

Norton resistance is obtained by a single DC test: deactivate all independent sources, apply a test voltage $V_t$ at the open terminals, and compute $R_N = V_t / I_t$. Because the entire computation fits within one simulation, the planner generates a single run.

\begin{tcolorbox}[sparcex,title={Planner output}]
{\ttfamily\scriptsize
num\_runs 1\\
run 1: Find the Norton equivalent resistance $R_N$ seen at terminals a--b.\\
Domain knowledge: Deactivate all independent sources (voltage sources shorted, current sources opened). Apply a 1\,V DC test source V\_test at terminals a--b. Measure I(V\_test). Norton resistance R\_N = 1\,V / I(V\_test).}
\end{tcolorbox}

\paragraph{P3: single-run using a built-in SPICE parameter sweep.}
\textbf{Q:} As $R_L$ varies from $5\,\Omega$ to $20\,\Omega$ in steps of $5\,\Omega$, would $V_L$ decrease or increase?

Rather than generating four separate runs, the planner recognises that a DC \texttt{.param} sweep can iterate over load values within a single simulation. This avoids redundant pipeline overhead and produces results at all four points simultaneously.

\begin{tcolorbox}[sparcex,title={Planner output}]
{\ttfamily\scriptsize
num\_runs 1\\
run 1: As $R_L$ varies from $5\,\Omega$ to $20\,\Omega$ in steps of $5\,\Omega$, would $V_L$ decrease or increase?\\
Domain knowledge: Declare RL as a .param and set the resistor to \{RL\}. Run a DC parameter sweep: RL from 5 to 20, step 5. Record V(load) at each of the four sweep points.}
\end{tcolorbox}

%----------------------------------------------------------------------
\subsubsection{Circuit Specification Agent}
%----------------------------------------------------------------------
This agent patches the \emph{circuit section} of~$\Po$, which contains component declarations and device models. 

\paragraph{C1: updating component values from question-supplied data.}
\textbf{Q:} Given $R{=}6\,\Omega$, source $u{=}12\sqrt{2}\sin(200t{+}30^\circ)$\,V, and $U_L{=}6\sqrt{2}$\,V, find the inductance $L$ in mH.

The base program~$\Po$ is auto-generated from the extracted netlist with placeholder values (typically 1) for all components. The circuit specification agent replaces those placeholders with the values stated in the question.

\begin{tcolorbox}[sparcex,title={Base (excerpt) -- placeholder values from netlist}]
{\ttfamily\scriptsize
R1 N1 N2 1\quad\quad (* placeholder -- will be updated to 6 *)\\
L1 N2 N3 1\quad\quad (* placeholder -- will be updated *)\\
V1 N1 0 AC 1\quad\quad (* amplitude and frequency not yet set *)}
\end{tcolorbox}
\begin{tcolorbox}[sparcex,title={Patch -- values from question applied}]
{\ttfamily\scriptsize
R1 N1 N2 6\quad\quad\quad\quad\quad\quad\quad\quad\quad (* R = 6 $\Omega$ from question *)\\
L1 N2 N3 15m\quad\quad\quad\quad\quad\quad\quad (* L $\approx$ 15\,mH, inferred *)\\
V1 N1 0 AC 16.97\quad\quad\quad\quad (* peak = 12$\sqrt{2}$ V *)\\
.param f = 200/(2*3.14159)\quad (* $\omega$=200 rad/s converted to Hz *)}
\end{tcolorbox}
The inductance $L$ is not given directly; it is inferred from the phasor relationship $U_L = \omega L I$ combined with the total current $I = U/\sqrt{R^2+(\omega L)^2}$. Solving yields $L \approx 15$\,mH, which the agent encodes directly.

\paragraph{C2: inserting a missing device model for a transistor.}
\textbf{Q:} Find the DC collector-emitter voltage $V_{CE}$ of the NPN transistor $Q_1$.

The netlist extracted from the diagram identifies $Q_1$ as an NPN device but does not supply the required \texttt{.model} statement, because model parameters are not visible in schematic diagrams. Without a \texttt{.model} entry, ngspice cannot determine the transistor's electrical characteristics and will fail with a model-not-found error. The circuit specification agent inserts a standard small-signal NPN model and rewrites the element line to reference it.

\begin{tcolorbox}[sparcex,title={Base (excerpt) -- model name absent}]
{\ttfamily\scriptsize
* BJT element line as extracted from netlist\\
Q1 NC NB NE\quad (* ngspice requires: Q<n> C B E <model\_name> *)\\
* No .model statement present -- will cause a fatal error}
\end{tcolorbox}
\begin{tcolorbox}[sparcex,title={Patch -- model added and element line completed}]
{\ttfamily\scriptsize
* Standard small-signal NPN model\\
.model NPN\_MOD NPN(Bf=200 Is=1e-14 Vaf=100)\\
* Element line updated to reference the model\\
Q1 NC NB NE NPN\_MOD}
\end{tcolorbox}

%----------------------------------------------------------------------
\subsubsection{Analysis Specification Agent}
%----------------------------------------------------------------------
This agent selects the appropriate ngspice analysis type based on the physical nature of the question, then rewrites the \emph{analysis section} of~$\Po$. It also adjusts source declarations to match the required analysis (e.g., converting a DC voltage source to an AC source for frequency-domain simulation). It does not modify component values or output directives.

\paragraph{A1: steady-state question $\Rightarrow$ DC operating-point analysis.}
\textbf{Q:} What is the steady-state output voltage $V_{\mathrm{out}}$ of the op-amp integrator circuit?

The keyword ``steady state'' with a constant input implies that all capacitors are fully charged and no current flows through them: a DC condition. The base program was initially configured for AC analysis (e.g., left over from a frequency-response template). The agent replaces the AC source with a DC source and swaps the \texttt{.ac} directive for \texttt{.op}.

\begin{tcolorbox}[sparcex,title={Base (excerpt) -- incorrectly set up for AC}]
{\ttfamily\scriptsize
V1 N1 0 AC 0.1\quad (* AC source -- wrong for steady-state query *)\\
.ac lin 1 1 1\quad\quad (* AC frequency sweep -- wrong analysis type *)}
\end{tcolorbox}
\begin{tcolorbox}[sparcex,title={Patch -- switched to DC operating point}]
{\ttfamily\scriptsize
V1 N1 0 DC 0.1\quad (* AC $\to$ DC: caps become open, inductors become short *)\\
.op\quad\quad\quad\quad\quad\quad\quad\quad (* replaces .ac; computes node voltages at DC steady state *)}
\end{tcolorbox}

\paragraph{A2: frequency-domain question $\Rightarrow$ AC small-signal analysis.}
\textbf{Q:} At $f{=}50$\,Hz, find the RMS voltage across $R_L$.

The phrase ``at $f{=}50$\,Hz'' identifies this as a small-signal AC query. The base program was configured for DC, so the agent converts the voltage source to AC type and replaces \texttt{.op} with a single-frequency \texttt{.ac} sweep.

\begin{tcolorbox}[sparcex,title={Base (excerpt) -- configured for DC}]
{\ttfamily\scriptsize
V1 N1 0 DC 220\quad (* DC source -- wrong for frequency-domain query *)\\
.op\quad\quad\quad\quad\quad\quad\quad\quad (* DC analysis -- no frequency information *)}
\end{tcolorbox}
\begin{tcolorbox}[sparcex,title={Patch -- switched to AC analysis at 50\,Hz}]
{\ttfamily\scriptsize
.param f = 50\quad\quad\quad\quad\quad\quad\quad\quad (* frequency in Hz *)\\
V1 N1 0 AC 220\quad\quad\quad\quad\quad\quad (* DC $\to$ AC source *)\\
.ac lin 1 \{f\} \{f\}\quad\quad\quad\quad (* single-frequency sweep; .op removed *)}
\end{tcolorbox}

\paragraph{A3: time-domain event question $\Rightarrow$ transient analysis.}
\textbf{Q:} After the switch opens at $t{=}0$, find the inductor current $i_L$ at $t{=}5$\,ms.

The question asks for a waveform value at a specific point in time, which requires a time-domain transient simulation. The base program uses a constant DC source, which cannot model a switching event. The agent replaces it with a PULSE source that drops from on to off at $t{=}0$, and sets a \texttt{.tran} analysis long enough to capture the $t{=}5$\,ms instant.

\begin{tcolorbox}[sparcex,title={Base (excerpt) -- constant DC, no switch model}]
{\ttfamily\scriptsize
V1 N1 0 DC 12\quad (* constant voltage; switch event not modelled *)\\
.op\quad\quad\quad\quad\quad\quad\quad\quad (* steady-state only; no time axis *)}
\end{tcolorbox}
\begin{tcolorbox}[sparcex,title={Patch -- PULSE source models switch opening at $t{=}0$}]
{\ttfamily\scriptsize
* PULSE(V\_on V\_off T\_delay T\_rise T\_fall T\_width T\_period)\\
V1 N1 0 PULSE(12 0 0 1n 1n 1 2)\\
\phantom{V1 N1 0 }\quad (* 12\,V $\to$ 0\,V at t=0, 1\,ns transition, stays off *)\\
.tran 10u 10m\quad (* step=10\,$\mu$s, stop=10\,ms; captures t=5\,ms *)}
\end{tcolorbox}

%----------------------------------------------------------------------
\subsubsection{Output Specification Agent}
%----------------------------------------------------------------------
This agent determines which quantities to extract from the simulator and emits the appropriate \texttt{.print} or \texttt{.meas} directives. 

\paragraph{O1: directly measurable branch quantities using \texttt{.print}.}
\textbf{Q:} Find the current $I_s$ through voltage source $V_1$ and the voltage $V_s$ across current source $I_1$.

Both quantities are natively accessible as simulator observables; a simple \texttt{.print} directive suffices.

\begin{tcolorbox}[sparcex,title={Patch applied}]
{\ttfamily\scriptsize
.print op I(V1)\quad\quad\quad (* branch current through V1 *)\\
.print op V(N3,N2)\quad (* voltage across I1, from N3 to N2 *)}
\end{tcolorbox}

\paragraph{O2: derived DC quantity via \texttt{.meas PARAM}.}
\textbf{Q:} Find the collector-emitter voltage $V_{CE}$ of transistor $Q_1$. Since $V_{CE}$ is the difference between the collector and emitter node voltages, the agent constructs a \texttt{.measure PARAM} expression to compute it arithmetically.

\begin{tcolorbox}[sparcex,title={Patch applied}]
{\ttfamily\scriptsize
.measure op Vce PARAM='V(NC)-V(NE)'\\}
\end{tcolorbox}

%----------------------------------------------------------------------
\subsubsection{Answer Generation Agent}
%----------------------------------------------------------------------
This agent receives the completed simulation log(s) and uses tool-based arithmetic to derive the final answer.

\paragraph{G1: reading a simulation result to resolve a multiple-choice question.}
\textbf{Q:} Will $V_{\mathrm{out}}$ saturate to $+V_{DD}$ (A), $-V_{EE}$ (B), equal $0.1$\,V (C), or $-0.1$\,V (D)?

The simulation computes the op-amp output node voltage under DC operating-point conditions. The agent inspects the magnitude and sign of the result to identify which answer choice it corresponds to.

\begin{tcolorbox}[sparcex,title={Simulation log (excerpt)}]
{\ttfamily\scriptsize
** DC operating point simulation results **\\
V(Nout) = -1.00000e+05\quad (* $-100{,}000$\,V -- below any physical supply rail *)}
\end{tcolorbox}
The value is a very large negative number, which is the simulator's representation of negative rail saturation (the ideal op-amp model drives the output to $-\infty$ when it saturates; in practice the output clamps to $-V_{EE}$). Options C and D are ruled out because the simulated magnitude is orders of magnitude larger than $0.1$\,V. \textbf{Answer: B}.

\paragraph{G2: tool-assisted derivation from simulation-confirmed parameters.}
\textbf{Q:} After the switch opens, when does $Q_1$ leave the active region? (A)~10\,ms \enspace (B)~25\,ms \enspace (C)~100\,ms \enspace (D)~50\,ms

Run~1 (switch closed, DC) confirms $R_1 = 4300\,\Omega$, $C_1 = 5\,\mu$F, and the initial collector voltage $V_0 = 5$\,V. $Q_1$ leaves the active region when $V_{CE}$ decays to $V_{\mathrm{sat}} \approx 0.7$\,V. The agent delegates both arithmetic steps to the calculator tool rather than computing them inline.

\begin{tcolorbox}[sparcex,title={Tool-assisted derivation (from confirmed simulation parameters)}]
{\scriptsize
Step 1 -- compute time constant:\quad\texttt{calc}($R_1 \times C_1$) $= 4300 \times 5\!\times\!10^{-6} = 21.5$\,ms\\[2pt]
Step 2 -- solve $V_0\,e^{-t/\tau} = V_{\mathrm{sat}}$:\quad
$t = \tau\ln(V_0/V_{\mathrm{sat}})$\\[2pt]
\quad\texttt{calc}($21.5\!\times\!10^{-3} \times \ln(5/0.7)$) $\approx 42.4$\,ms $\to$ nearest option is 50\,ms.}
\end{tcolorbox}
\textbf{Answer: D}.
%----------------------------------------------------------------------
\subsubsection{Error Diagnosis Agent}
%----------------------------------------------------------------------
When ngspice exits with an error, the error diagnosis agent reads the error message together with the complete edit history (all patches applied so far) and identifies which section of~$\Po$ is responsible. It then selectively re-invokes only the faulty specification agent, rather than restarting the entire pipeline from scratch. This targeted repair minimises redundant work on iterations where other sections are already correct.

\paragraph{E1: a spurious keyword displaces the required model name in a diode declaration.}
The ngspice diode element syntax is: \texttt{D<name> <pos\_node> <neg\_node> <model\_name>}. The circuit specification agent incorrectly inserted the keyword \texttt{DC} before the model name. ngspice consumed \texttt{DC} as the model name, then encountered \texttt{DIO} as an unexpected extra token and reported it as an unknown parameter.

\begin{tcolorbox}[sparcex,title={Failing circuit line and ngspice error}]
{\ttfamily\scriptsize
VD1 N\_VD1\_sense N3 DC DIO\quad (* ``DC'' must not appear here *)\\
Error: unknown parameter (dio)\quad (* ngspice parsed DC as model name *)}
\end{tcolorbox}
\begin{tcolorbox}[sparcex,title={Diagnosis and targeted fix}]
{\ttfamily\scriptsize
Root cause: ``DC'' is not valid in diode element syntax.\\
\phantom{Root cause: }ngspice reads positional fields: pos, neg, model\_name.\\
\phantom{Root cause: }``DC'' occupies the model\_name field, so ``DIO'' is extra.\\
Responsible section: circuit specification.\\
Re-invoke: circuit specification agent (only).\\
Fix: VD1 N\_VD1\_sense N3 DIO\quad (* remove ``DC''; DIO is now the model *)}
\end{tcolorbox}

\section{Related Work}\label{app:related work}

\paragraph{Electrical Circuit Analysis.}
To the best of our knowledge, \textit{no} prior work directly addresses question answering over electrical circuit diagrams that requires mathematically grounded reasoning. Existing efforts fall into two categories: (1) visual perception and extraction, such as circuit component recognition~\cite{CircuitVQA, ElectroVizQA} and netlist extraction~\cite{shi2025amsnet}, and (2) benchmarks without concrete mechanisms~\cite{li2025eee}.
One prior work~\cite{nau2025spiceassistant} does use SPICE but for a completely \textit{different} task: electrical power system design automation. 
\\\noindent\textbf{General Diagram Question Answering.}
Prior work on general diagram question answering falls into two categories. The first focuses on semantic interpretation via prompting mechanisms~\cite{wang2024cog, wang2025glfomr, wang2024charxiv, farahani2025chart} or external knowledge sources~\cite{notemr, reflectiva}. In contrast, \systemname~targets mathematically grounded reasoning that requires multi-step quantitative analysis rather than semantic interpretation.
The second line of work explores tool-augmented reasoning by converting diagrams into structured symbolic representations~\cite{bauer2023neuro, suri2025follow} or by synthesizing and executing programs for quantitative reasoning~\cite{plotqa, das2024mathsensei, toolvqa}. While these approaches also generate executable programs, they primarily rely on general-purpose code (e.g., Python) that requires explicitly \textit{formulating} the governing equations and computation logic. In contrast, \systemname{} synthesizes SPICE programs, which encode the circuit structure and analysis configuration and \textit{delegate} equation selection and numerical solving to a physics-based simulator. Our work is most closely related to this line of research, but differs in specializing program synthesis to simulator-native code, enabling physics-grounded, multi-step reasoning over circuit diagrams that prior approaches do not support.

\section{Dataset Annotation Process}
In this section, we describe the annotation procedure used to construct our datasets.

\subsection{\eeebench{}}
\label{app:annotation_errors}
For \eeebench{}, we first prompt GPT-5 to extract the circuit structure in netlist format and then manually verify and correct the extracted netlists. Below, we summarize the error types that required human intervention, as well as the image categories that were excluded and the reasons for their exclusion.

\subsubsection{Annotation Error Categories}

During image annotation, we observed several recurring error patterns that required human correction. These errors can be grouped into three high level categories.

\paragraph{Component Semantics Errors.}
These errors arise when components are detected but assigned incorrect or incomplete semantic attributes. Common cases include incorrect polarity assignment for diodes and capacitors, although polarity detection for voltage and current sources was generally accurate. Multi terminal components were frequently misrepresented: bipolar junction transistors were often ignored or annotated with incorrect terminal counts or ordering; MOSFETs were detected more reliably but lacked device type information and were reduced to two terminal elements; three terminal regulators were annotated with missing pins; and timer ICs such as the 555 were not annotated at all. Several components were systematically misclassified, including voltmeters being interpreted as voltage sources and symbol variants such as Zener diodes, thyristors, photodiodes, rheostats, and controlled sources being collapsed into simpler component types despite inline textual labels indicating their intended function.

\paragraph{Structural and Connectivity Errors.}
This category captures failures in constructing a correct circuit graph. In complex circuits with multiple active devices and dense wiring, the model frequently failed to detect wire intersections, leading to incorrect node assignments or reuse of node identifiers across disconnected regions. Parallel branches with current labels were sometimes ambiguously represented, making it unclear which branch a measurement referred to. Additional issues include failure to represent shorted terminals explicitly, incorrect handling of multi position switches such as SPDT configurations, and inability to separate multiple circuits or answer options presented within a single image. These errors were corrected through manual node reassignment, explicit inline comments, or circuit segmentation.

\paragraph{Labeling and Parameterization Errors.}
These errors involve missing or inconsistent numerical and symbolic information rather than structural faults. Common examples include missing component values or units, confusion between voltage source values and voltage measurement labels, omission of control parameters such as switch duty cycles, and inconsistent naming of components with identical values. When multiple components shared the same nominal value, unique identifiers were manually assigned while preserving value equality. When label ambiguity could not be resolved structurally, clarifying inline comments were added.

\subsubsection{Excluded Image Categories}

Certain image types were excluded from annotation because they cannot be faithfully represented using schematic level circuit annotations or netlist style abstractions.

\paragraph{Non Circuit Figures.}
Images containing no circuits, such as component photographs, charts, plots, oscilloscope outputs, and purely textual figures, were excluded.

\paragraph{Logic and Discrete Abstractions.}
Logic circuits, Karnaugh maps, truth tables, state diagrams, and assembly level programs were omitted, as they require symbolic or temporal representations beyond electrical schematics.

\paragraph{System Level and Physical Diagrams.}
System block diagrams, signal flow graphs, frequency response plots, magnetic circuits, semiconductor band diagrams, and diagrams requiring physical parameters such as wire cross section or length were excluded because they do not admit a direct translation into executable \texttt{ngspice} netlists and cannot be simulated within the circuit level modeling assumptions supported by \texttt{ngspice}.

\paragraph{High Complexity Power and Machine Systems.}
Diagrams involving motors, generators, transformers, transmission lines, power plants, and power system networks were excluded because they rely on large scale system models, distributed parameters, or domain specific abstractions that cannot be faithfully represented or executed using standard \texttt{ngspice} schematic annotations.

\subsection{\amsnet{}}
\label{app:annotation_amsnet}
We predefine four circuit analysis tasks, including DC operating-point analysis (Prompt~\ref{appendix:prompt-analysis}), AC small-signal analysis (Prompt~\ref{appendix:prompt-analysis}), DC parameter sweep (Prompt~\ref{appendix:prompt-analysis}), and transient analysis (Prompt~\ref{appendix:prompt-analysis}), which cover common forms of reasoning required in circuit simulation. Each task is paired with a structured system prompt that specifies how to modify a netlist to enable the corresponding analysis and how to extract target quantities from simulation outputs.
Concretely, we first sample a circuit instance from the AMSNet dataset and randomly assign one of the predefined problem types. 
The circuit program and problem specification are then provided to an LLM agent with a system prompt shown in Fig.~\ref{appendix:prompt-analysis} with the corresponding analysis demonstration appended. 
The agent then iteratively edits the netlist and executes simulations.
Based on the resulting simulation outputs, the agent reasons about the circuit behavior and generates a corresponding natural-language question together with its numerical answer. Finally, we store the modified program along with the generated question-answer pair. 
Not to mention that, given an arbitrary circuit, not every circuit analysis will apply, which will result in simulation failure or not meaningful values. 
We provided detailed demonstrations for these cases in the system prompts, and the agents are instructed to output \texttt{Not Applicable} and its reasoning.

\section{Artifact Use}\label{appendix:artifact}
\subsection{Dataset License Information}
In accordance with ACL guidelines, we disclose the licenses of all datasets used in this work. We augment two existing datasets. \textbf{EEE-Bench} \cite{li2025eee} is released under the MIT License, which permits reuse, modification, and redistribution. \textbf{AMSNet} \cite{shi2025amsnet} is released under the GNU General Public License v3.0 (GPLv3), which allows use and modification under the condition that derivative works are distributed under the same license. Our augmentations preserve the original licensing terms of each dataset. Additionally, we will release our augmented datasets publicly under a GNU General Public License v3.0 (GPLv3). Because our datasets consist solely of electrical circuit data, they contain no personally identifiable information.

\subsection{Software and Multimodal Models}
We used the \texttt{ngspice} circuit simulator, which is released under the BSD-3-Clause license, permitting free use, modification, and redistribution.

The multimodal language models employed are publicly available and used under their respective licenses or terms of service:
\begin{itemize}
    \item \textbf{\textsc{GPT-5.1}} and \textbf{\textsc{GPT-4o}}: Proprietary model accessed via OpenAI API under OpenAI's Terms of Service.
    \item \textbf{\textsc{Claude Sonnet 4}}: Proprietary model accessed via Anthropic's API under Anthropic's terms of service.
    \item \textbf{\textsc{Qwen3-VL-32B Instruct}}: Released under the Apache-2.0 license.
    \item \textbf{\textsc{GLM-4.5V}}: Released under the MIT License on Hugging Face.
\end{itemize}

\section{Prompts}\label{app:prompts}
\subsection{Baselines}
The prompts used for the chain-of-thought (\textsc{CoT}) and code generation (\textsc{Code}) baselines are shown in Figures~\ref{appendix:prompt-cot} and~\ref{appendix:prompt-codegen}, respectively.

\begin{figure*}[!t]
\begin{lstlisting}[
  basicstyle=\footnotesize\normalfont,
  breaklines=true,
  frame=single,
  numbers=none,
  breakindent=0pt,
  backgroundcolor=\color{gray!10},
  aboveskip=4pt,
  belowskip=4pt,
  xleftmargin=6pt,
  xrightmargin=6pt,
  lineskip=0pt,
  columns=fullflexible
]
You are an NGSpice-based exam-question generator and solver. You will be given an NGSpice program as input. Your task is to generate ONE exam-style question and its correct answer by minimally editing the netlist, running a SPICE simulation, and reasoning about the results.

You MUST follow the workflow, constraints, and output format exactly.

OBJECTIVE
- Perform DC operating-point (OP) analysis on the given circuit
- Generate exam questions that require interpreting DC voltages and currents
- Answers must be obtained from simulation, not estimation

REQUIRED WORKFLOW
1) Read the given SPICE netlist.
2) Apply ONLY minimal edits needed to support DC operating-point analysis:
   - You may add .OP and .PRINT statements.
   - You may fix trivial syntax issues (e.g., missing units like "5V" -> "5").
   - Do NOT change circuit topology or add components unless strictly required for convergence.
3) Run the SPICE simulation.
4) Collect numeric results directly from the simulator output.
5) Generate:
   - Exam questions
   - Correct answers
   - Brief reasoning grounded in the simulation results

You must run the simulation before producing questions and answers.

PRINTING REQUIREMENTS
- If node 1 exists, print V(1).
- If node 2 exists, print V(2).
- If a voltage source named V1 exists, print I(V1).

If these exact nodes or sources do not exist:
- Print the closest reasonable equivalents.
- Prefer node voltages and currents through independent voltage sources.

REASONING REQUIREMENTS
- Provide 1-3 sentences interpreting the DC operating-point results.
- Interpretation must be based on the numeric output.
- Examples of acceptable reasoning:
  - Whether a transistor is on or off
  - Whether the current is zero or non-zero
  - Explanation of current sign conventions
\end{lstlisting}
\caption{System prompt for \amsnet{} annotation. We randomly sample a different demonstration for a different analysis type and appended it to the system prompt.}
\label{appendix:prompt-analysis-system}
\end{figure*}

\begin{figure*}[!t]
\begin{lstlisting}[
  basicstyle=\footnotesize\normalfont,
  breaklines=true,
  frame=single,
  numbers=none,
  breakindent=0pt,
  backgroundcolor=\color{gray!10},
  aboveskip=4pt,
  belowskip=4pt,
  xleftmargin=6pt,
  xrightmargin=6pt,
  lineskip=0pt,
  columns=fullflexible
]
FAILURE HANDLING
If the DC operating-point analysis is not applicable or produces unusable output
(e.g., simulator errors, non-convergence, floating nodes, or required quantities cannot be printed),
output EXACTLY:

NOT_APPLICABLE: <one short, concrete reason>

Then STOP. Do not generate questions, answers, or reasoning.

EXAMPLE
Given program
* SPICE Netlist for circuit 0
M1 1 2 0 0 NMOS W=1u L=1u
V1 2 0 5
.MODEL NMOS NMOS (LEVEL=1 VTO=1 KP=1.0e-4 LAMBDA=0.02)
.END

Edits
Add DC operating-point analysis and print node voltages and source current.

.OP
.PRINT OP V(1) V(2) I(V1)

Final program
* SPICE Netlist for circuit 0
M1 1 2 0 0 NMOS W=1u L=1u
V1 2 0 5
.MODEL NMOS NMOS (LEVEL=1 VTO=1 KP=1.0e-4 LAMBDA=0.02)

.OP
.PRINT OP V(1) V(2) I(V1)
.END

Simulation results
V(1) = 4.21 V
V(2) = 5.00 V
I(V1) = -0.78 mA

Questions
1. What are the voltages at nodes 1 and 2 at the DC operating point?
2. What is the DC current drawn from the voltage source V1?

Answers
1. V(1) = 4.21 V, V(2) = 5.00 V
2. I(V1) = -0.78 mA
\end{lstlisting}
\caption{DC operating-point analysis demonstration prompt.}
\label{appendix:prompt-dc-analysis}
\end{figure*}

\begin{figure*}[!t]
\begin{lstlisting}[
  basicstyle=\footnotesize\normalfont,
  breaklines=true,
  frame=single,
  numbers=none,
  breakindent=0pt,
  backgroundcolor=\color{gray!10},
  aboveskip=4pt,
  belowskip=4pt,
  xleftmargin=6pt,
  xrightmargin=6pt,
  lineskip=0pt,
  columns=fullflexible,
  mathescape
]
FAILURE HANDLING
If AC analysis is not applicable or produces unusable output, output EXACTLY:

NOT_APPLICABLE: <one short, concrete reason>

Use NOT_APPLICABLE for cases including (but not limited to):
- No independent source exists to attach an AC magnitude
- Simulator errors or non-convergence
- All printed AC quantities are identically zero across the sweep (magnitude and phase not meaningful)
- The requested transfer quantity is undefined (e.g., division by zero)

Then STOP. Do not generate questions, answers, or reasoning.

EXAMPLE
Given program
* SPICE Netlist for circuit 0
M1 1 2 0 0 NMOS W=1u L=1u
V1 2 0 5
.MODEL NMOS NMOS (LEVEL=1 VTO=1 KP=1.0e-4 LAMBDA=0.02)
.END

Edits
Add AC magnitude to the source, add AC sweep, and print AC voltages and source current.

Change source to include AC excitation:
V1 2 0 DC 5 AC 1

Add AC analysis and printing:
.AC DEC 10 1k 1G
.PRINT AC V(1) V(2) I(V1)

Final program
* SPICE Netlist for circuit 0
M1 1 2 0 0 NMOS W=1u L=1u
V1 2 0 DC 5 AC 1
.MODEL NMOS NMOS (LEVEL=1 VTO=1 KP=1.0e-4 LAMBDA=0.02)

.AC DEC 10 1k 1G
.PRINT AC V(1) V(2) I(V1)
.END

Simulation results
At f = 1.000000e+06 Hz:
V(1) = 0.000000e+00, 0.000000e+00
V(2) = 1.000000e+00, 0.000000e+00
I(V1) = -1.000000e-04, 0.000000e+00

Questions
1. At 1 MHz, what are the magnitude and phase of V(2)?
2. At 1 MHz, what are the magnitude and phase of I(V1)?

Answers
1. |V(2)| = 1.0, angle V(2) = 0 deg
2. |I(V1)| = 1.0e-04 A, angle I(V1) = 0 deg
\end{lstlisting}
\caption{AC small-signal analysis demonstration prompt.}
\label{appendix:prompt-ac-analysis}
\end{figure*}

\begin{figure*}[!t]
\begin{lstlisting}[
  basicstyle=\footnotesize\normalfont,
  breaklines=true,
  frame=single,
  numbers=none,
  breakindent=0pt,
  backgroundcolor=\color{gray!10},
  aboveskip=4pt,
  belowskip=4pt,
  xleftmargin=6pt,
  xrightmargin=6pt,
  lineskip=0pt,
  columns=fullflexible
]
FAILURE HANDLING
If DC sweep analysis is not applicable or produces unusable output
(e.g., no voltage source to sweep, simulator errors, non-convergence, or required quantities cannot be printed),
output EXACTLY:

NOT_APPLICABLE: <one short, concrete reason>

Then STOP. Do not generate questions, answers, or reasoning.

EXAMPLE
Given program
* SPICE Netlist for circuit 0
M1 1 2 0 0 NMOS W=1u L=1u
V1 2 0 5
.MODEL NMOS NMOS (LEVEL=1 VTO=1 KP=1.0e-4 LAMBDA=0.02)
.END

Edits
Add DC sweep of the voltage source from 0 V to 5 V in 0.5 V steps and print node voltage and source current.

.DC V1 0 5 0.5
.PRINT DC V(2) I(V1)

Final program
* SPICE Netlist for circuit 0
M1 1 2 0 0 NMOS W=1u L=1u
V1 2 0 5
.MODEL NMOS NMOS (LEVEL=1 VTO=1 KP=1.0e-4 LAMBDA=0.02)

.DC V1 0 5 0.5
.PRINT DC V(2) I(V1)
.END

Simulation results
V1 = 0.0 V   I(V1) = 0.0 A
V1 = 0.5 V   I(V1) = 0.0 A
V1 = 1.0 V   I(V1) = 0.0 A
V1 = 1.5 V   I(V1) = -0.10 mA
V1 = 2.0 V   I(V1) = -0.25 mA
V1 = 2.5 V   I(V1) = -0.45 mA
...

Questions
1. At what input voltage does the current through V1 first become non-zero?
2. What is the current through V1 when the swept voltage is 2.0 V?

Answers
1. The current first becomes non-zero at approximately 1.5 V.
2. At 2.0 V, I(V1) = -0.25 mA.
\end{lstlisting}
\caption{DC parameter sweep demonstration prompt.}
\label{appendix:prompt-dc-sweep-analysis}
\end{figure*}

\begin{figure*}[!t]
\begin{lstlisting}[
  basicstyle=\footnotesize\normalfont,
  breaklines=true,
  frame=single,
  numbers=none,
  breakindent=0pt,
  backgroundcolor=\color{gray!10},
  aboveskip=4pt,
  belowskip=4pt,
  xleftmargin=6pt,
  xrightmargin=6pt,
  lineskip=0pt,
  columns=fullflexible
]
FAILURE HANDLING
If transient analysis is not applicable or produces unusable output
(e.g., simulator errors, non-convergence, floating nodes, or required quantities cannot be printed),
output EXACTLY:

NOT_APPLICABLE: <one short, concrete reason>

Then STOP. Do not generate questions, answers, or reasoning.

EXAMPLE
Given program
* SPICE Netlist for circuit 0
M1 1 2 0 0 NMOS W=1u L=1u
V1 2 0 5
.MODEL NMOS NMOS (LEVEL=1 VTO=1 KP=1.0e-4 LAMBDA=0.02)
.END

Edits
Add transient analysis for 100 ns with 1 ns time step and print voltages and source current.

.TRAN 1n 100n
.PRINT TRAN V(1) V(2) I(V1)

Final program
* SPICE Netlist for circuit 0
M1 1 2 0 0 NMOS W=1u L=1u
V1 2 0 5
.MODEL NMOS NMOS (LEVEL=1 VTO=1 KP=1.0e-4 LAMBDA=0.02)

.TRAN 1n 100n
.PRINT TRAN V(1) V(2) I(V1)
.END

Simulation results
At t = 100 ns:
V(1) = 4.21 V
V(2) = 5.00 V
I(V1) = -0.78 mA

Questions
1. At t = 100 ns, what are the voltages at nodes 1 and 2?
2. At t = 100 ns, what is the current through the voltage source V1?

Answers
1. V(1) = 4.21 V, V(2) = 5.00 V
2. I(V1) = -0.78 mA
\end{lstlisting}
\caption{transient analysis demonstration prompt.}
\label{appendix:prompt-transient-analysis}
\end{figure*}

\begin{figure*}[!t]
\begin{lstlisting}[
  basicstyle=\footnotesize\normalfont,
  breaklines=true,
  frame=single,
  numbers=none,
  breakindent=0pt,
  backgroundcolor=\color{gray!10},
  aboveskip=4pt,
  belowskip=4pt,
  xleftmargin=6pt,
  xrightmargin=6pt,
  lineskip=0pt,
  columns=fullflexible
]
You are an expert in circuit analysis.

Given the provided inputs, solve the circuit problem using step by step reasoning.

Requirements:

1. Reason carefully using the circuit diagram, schema, and question.
2. Apply appropriate circuit laws and mathematical reasoning.
3. If the question is multiple choice, output only the letter of the correct option.
4. Provide the final answer in the following format:

<final_answer>ANSWER</final_answer>

Input will be provided as follows.

[Schema only setting]
Schema: {schema}
Question: {question}

[Diagram + schema setting]
Diagram: <image>
Schema: {schema}
Question: {question}

[Diagram only setting]
Diagram: <image>
Question: {question}

Output only the reasoning process and the final answer.
\end{lstlisting}
\caption{Prompt used for the chain-of-thought (\textsc{CoT}) baselines}
\label{appendix:prompt-cot}
\end{figure*}

\begin{figure*}[!t]
\begin{lstlisting}[
  basicstyle=\footnotesize\normalfont,
  breaklines=true,
  frame=single,
  numbers=none,
  breakindent=0pt,
  backgroundcolor=\color{gray!10},
  aboveskip=4pt,
  belowskip=4pt,
  xleftmargin=6pt,
  xrightmargin=6pt,
  lineskip=0pt,
  columns=fullflexible
]
You are an expert in circuit analysis and symbolic mathematics.

Given a circuit schema and a natural language question, generate executable Python code using SymPy to solve the problem symbolically.

Your code should:

1. Define all required symbolic variables.
2. Construct equations using circuit laws such as Ohm's law, Kirchhoff's Voltage Law (KVL), and Kirchhoff's Current Law (KCL).
3. Solve the resulting system of equations symbolically or numerically.
4. Print the final answer.

Requirements:

1. Include all necessary imports.
2. Generate complete executable Python code only.
3. Wrap the code between ```python and ``` markers.
4. Print the final result using:

print(f"FINAL_ANSWER: {answer}")

Input will be provided as follows.

Schema: {schema}

Question: {question}

Output only the generated SymPy code.
\end{lstlisting}
\caption{Prompt used for the \textsc{Code} baseline}
\label{appendix:prompt-codegen}
\end{figure*}
\subsection{Planner Agent}
The prompt used by the planner agent is shown in Figure~\ref{appendix:prompt-planner}.

\begin{figure*}[!t]
\begin{lstlisting}[
  basicstyle=\footnotesize\normalfont,
  breaklines=true,
  frame=single,
  numbers=none,
  breakindent=0pt,
  backgroundcolor=\color{gray!10},
  aboveskip=4pt,
  belowskip=4pt,
  xleftmargin=6pt,
  xrightmargin=6pt,
  lineskip=0pt,
  columns=fullflexible
]
You are an expert in planning NGSpice simulations from a circuit schema and a natural language question.

Given the user's question, the circuit schema, and relevant domain knowledge, your job is to decide how many NGSpice simulations are required and how the question should be instantiated for each run.

Your task is as follows.

1. If the question asks about a range, change, maximum, or minimum of input values, generate multiple simulations with concrete input values.

2. If the question asks about a range, change, maximum, or minimum of output values, decide whether multiple simulations are needed to capture that variation, and generate them if required.

3. If the question describes a pre switch and post switch scenario, generate two simulations: one for the circuit before the switch and one for the circuit after the switch. Rephrase the question for each run to reflect the corresponding circuit state.

4. If the requested quantity can be obtained using a single NGSpice sweep, such as a DC sweep, parameter sweep, or AC frequency sweep, generate only one simulation and keep the original question unchanged.

5. When multiple simulations are required, clearly specify the number of runs and provide a rephrased question for each run. Preserve all fixed values from the original question and vary only the quantities that are implied to change.

6. Limit the total number of runs to at most 5.

The output format must be followed exactly.

num_runs X
run 1: rephrased question for run 1
run 2: rephrased question for run 2
...
run X: rephrased question for run X

Input will be provided as follows.

Question: {question}
Schema: {schema}
Domain knowledge: {dk}

Output only the number of runs and the rephrased questions in the specified format.
\end{lstlisting}
\caption{Prompt used for the planner agent}
\label{appendix:prompt-planner}
\end{figure*}
\subsection{Circuit Specification Agent}
The prompt used by the circuit specification agent is shown in Fig.~\ref{appendix:prompt-value-model}.

\begin{figure*}[!t]
\begin{lstlisting}[
  basicstyle=\footnotesize\normalfont,
  breaklines=true,
  frame=single,
  numbers=none,
  breakindent=0pt,
  backgroundcolor=\color{gray!10},
  aboveskip=4pt,
  belowskip=4pt,
  xleftmargin=6pt,
  xrightmargin=6pt,
  lineskip=0pt,
  columns=fullflexible
]
You are an expert in electrical engineering and NGSpice. Your job is to:
1. Update component values in the netlist based on the question requirements
2. Add or correct any missing .model statements for devices that require them

## PART 1: Updating Values

### Allowed Value Edits

#### 1.1 Update numeric values of existing elements or sources
You may change a numeric literal or parameter expression **only if**:
- The question explicitly gives a value (e.g., R1 = 2 kOhm, V1 = 10 V, C3 = 4 uF), AND
- The name in the question **exactly matches** an element name in the netlist (e.g., R1, V1, Rload).

#### 1.2 Update models of elements when specific parameters are given
For example, if the question specifies a Zener diode with a specific Zener voltage:
.model DZ D(Is=1e-15 N=1 BV=5)

#### 1.3 Introduce frequency parameters when needed
If the question gives a frequency:
- For **Hz**: .param f = <value>
- For **rad/s**: .param f = <value>/(2*3.14159265)

When modifying an AC analysis line, use the parameter:
.ac lin 1 {f} {f}

#### 1.4 Element names must remain type-correct
In NGSpice, device names must begin with a letter corresponding to their device type:
- V -> voltage source  
- I -> current source  
- R -> resistor  
- C -> capacitor  
- L -> inductor  
- D -> diode  
- Q / M -> transistors  

If the question specifies a type that conflicts with the current name, you must **rename** the element.

#### 1.5 Model open circuit elements
If the question specifies an open circuit (e.g., an open resistor), you will set its value to an extremely large number (e.g., 1e12).
#### 1.6 Type mismatch in schema
If the schema indicates one type but uses another (e.g, a voltage source with the type NPN), correct the type to match the actual device.

### Forbidden Value Edits (do not do any of these):
- Do not add or remove any component or source unless explicitly required.
- Do not change values that are not explicitly given in the question.
---

## PART 2: Adding Missing Models
Only the following require models:
- Diodes: D
- BJTs: Q
- MOSFETs: M
- JFETs: J

### 2.1 General `.model` syntax

    .model <model_name> <device_type> (<parameters>)

Examples:
...

edit:

"""
\end{lstlisting}
\caption{Prompt used for circuit specification agent}
\label{appendix:prompt-value-model}
\end{figure*}

\subsection{Analysis Specification Agent}
The prompt used by the analysis specification agent is shown in Figure~\ref{appendix:prompt-analysis}.

\begin{figure*}[!t]
\begin{lstlisting}[
  basicstyle=\footnotesize\normalfont,
  breaklines=true,
  frame=single,
  numbers=none,
  breakindent=0pt,
  backgroundcolor=\color{gray!10},
  aboveskip=4pt,
  belowskip=4pt,
  xleftmargin=6pt,
  xrightmargin=6pt,
  lineskip=0pt,
  columns=fullflexible
]
You are an expert in electrical circuit analysis. Given a circuit schema, a question, and domain knowledge, your job is to:

1. Select the correct NGSpice analysis type (DC or AC).
2. Produce a single NGSpice edit specification that configures the netlist accordingly.

## Analysis Type Guidelines

Choose **DC analysis** for steady state behavior, including operating point and bias, comparator output, rail saturation or clipping, DC gain or offset, device conduction states, and responses to constant inputs.

Choose **AC analysis** for frequency dependent small signal behavior, including frequency response, gain or phase at a specified frequency, Bode quantities, cutoff or bandwidth, resonance, impedance at a frequency, and any query stated at ( f ) or ( \omega ).

## DC Analysis Instructions

If DC analysis is needed:

* Convert AC sources to DC sources.
* Set analysis to `.op`.
* Delete any `.ac` or `.tran` statements.
* Remove `.print` and `.save` statements that refer to AC or transient analysis.

## AC Analysis Instructions

If AC analysis is needed:

1. Ensure at least one AC stimulus exists by converting a relevant source to:
   `Vx <pos> <neg> AC <value>` or `Ix <pos> <neg> AC <value>`
2. Set analysis to a valid `.ac` statement:

   * If the question specifies ( f ): use `.ac lin 1 {f} {f}`
   * If the question gives ( \omega ): set `.param f = omega/(2*3.14159265)`
3. Delete any `.op` or `.tran` statements unless explicitly required.
4. Replace DC or transient outputs with AC outputs, for example:
   `.print ac V(node) I(source)`

## Output Format

Return three parts in order:

1. **Analysis Type**: DC or AC, with a brief justification.
2. **Reasoning**: Step by step edits implied by the chosen analysis.
3. **Edit Specification**: The NGSpice edit specification beginning with `edit:`.
\end{lstlisting}
\caption{Prompt used for analysis specification agent}
\label{appendix:prompt-analysis}
\end{figure*}

\subsection{Output Specification Agent}
The prompt used by the output specification agent is shown in Fig.~\ref{appendix:prompt-output}.

\begin{figure*}[!t]
\begin{lstlisting}[
  basicstyle=\footnotesize\normalfont,
  breaklines=true,
  frame=single,
  numbers=none,
  breakindent=0pt,
  backgroundcolor=\color{gray!10},
  aboveskip=4pt,
  belowskip=4pt,
  xleftmargin=6pt,
  xrightmargin=6pt,
  lineskip=0pt,
  columns=fullflexible,
  upquote=true
]
### NGSpice `.measure` Usage Guide

The `.measure` (or `.meas`) statement instructs NGSpice to compute a **scalar value** from simulation results, such as time, voltage, current, power, peaks, averages, integrals, or event times.

**General form**

.measure <analysis> <name> <type> <conditions...>

where `<analysis>` in {`tran`, `ac`, `dc`} and `<type>` in {`WHEN`, `MAX`, `MIN`, `AVG`, `INTEG`, `PARAM`}.

### Measurement Types

**WHEN (event time)**

.measure tran t1 WHEN V(node)=value
.measure tran t2 WHEN I(L1)=0 CROSS=2

Returns the time of the specified event. Optional qualifiers: `CROSS=n`, `RISE=n`, `FALL=n`.

.measure tran vpp PARAM='vmax - vmin'

### AC Power Factor and Phase
For power factor, phase angle, or leading/lagging behavior, measure source voltage and current:
.measure ac Vrms RMS V(pos,neg)
.measure ac Irms RMS I(Vsrc)
.measure ac Pavg AVG (V(pos,neg)*I(Vsrc))
.measure ac pf PARAM='Pavg/(Vrms*Irms)'
Use a single frequency point (e.g., `.ac lin 1 {f} {f}`).

### DC Operating Point Quantities
For DC quantities such as ( V_{CE} ) or Q point:
.measure op vce PARAM='V(C)-V(E)'
.measure op ic  PARAM='-I(VCC)'
For "voltage across a load", always measure `V(load_pos) - V(load_neg)`.

### Task Instruction
Given a question and an NGSpice program, decide whether a `.measure` statement is required to answer the question.

* If required, output the appropriate `.measure` statement.
* If not required, explicitly state that no measurement is needed.
\end{lstlisting}
\caption{Prompt used for output specification agent}
\label{appendix:prompt-output}
\end{figure*}

\subsection{Answer Generation agent}
The prompt used by the answer generation agent is shown in Figure~\ref{appendix:prompt-answer}.

\begin{figure*}[!t]
\begin{lstlisting}[
  basicstyle=\footnotesize\normalfont,
  breaklines=true,
  frame=single,
  numbers=none,
  breakindent=0pt,
  backgroundcolor=\color{gray!10},
  aboveskip=4pt,
  belowskip=4pt,
  xleftmargin=6pt,
  xrightmargin=6pt,
  lineskip=0pt,
  columns=fullflexible
]
You are an expert in electrical circuit analysis and in interpreting NGSpice simulation outputs.

You will receive a circuit schema and diagram, one or more questions, an NGSpice program and its text output, a final question to answer, a reference reasoning trace, and additional domain knowledge.

Your task is to answer the final question using the NGSpice output as the primary source of truth.

Process:

* Interpret the NGSpice output and identify quantities relevant to the question.
* Use NGSpice results even if warnings are present.
* Derive the requested quantity step by step from reported voltages, currents, or phasors.
* Do not compute numeric values manually. All numeric calculations must be performed using the provided tools.
* For equivalent resistance or impedance, always compute it from measured voltage and current in the NGSpice output.
* Preserve polarity and node ordering exactly as stated in the question.
* For AC quantities, use complex values and compute magnitude or phase when required.
* If the required measurement is missing from the output, state that it cannot be determined.
* If comparing two cases, both cases must be present in the output.
* For multiple choice questions, verify each option numerically using the tools and select the correct one.

Output format:

* Show the sequence of tool calls.
* Give a brief explanation if needed.
* Clearly state the final answer.
* If an option letter is requested, output only the letter.
\end{lstlisting}
\caption{Prompt used for answer generation agent}
\label{appendix:prompt-answer}
\end{figure*}

\end{document}